\documentclass[journal,twoside,web]{ieeecolor}
\pdfoutput=1
\usepackage{tmi}
\usepackage{cite}
\usepackage{amsmath,amssymb,amsfonts}
\usepackage{algorithmic}
\usepackage{graphicx}
\usepackage{textcomp}
\usepackage{multirow}
\usepackage{multicol}
\usepackage{booktabs}
\usepackage{hyperref}
\usepackage{makecell} 
\usepackage{tabu}
\usepackage{bm}
\usepackage{threeparttable}
\usepackage[table]{xcolor}
\usepackage{subcaption} %
\usepackage{tikz} %
\usepackage{xcolor} %
\def\BibTeX{{\rm B\kern-.05em{\sc i\kern-.025em b}\kern-.08em
    T\kern-.1667em\lower.7ex\hbox{E}\kern-.125emX}}
\begin{document}
\definecolor{lightgreen}{RGB}{146, 208, 80}

\title{A Medical Multimodal Large Language Model for Pediatric Pneumonia}

\author{Weiwei Tian, Xinyu Huang, Tianhao Cheng, Wen He, Jinwu Fang, Rui Feng, Daoying Geng, Xiaobo Zhang
\vspace{-1.5em}
\thanks{Manuscript received September 4, 2024. This work was supported in part by the National Natural Science Foundation of China~(No.62172101), and in part by the Science and Technology Commission of Shanghai Municipality~(No.22511106003, No.23511100602) and Municipal Hospital Frontier Joint Research Project~(No.SHDC12024136), which studies on Evaluation Indicator Construction and Clinical Application Management for Diagnostic and Treatment Assistant Large-scale Model for Pediatric Severe Pneumonia. \textit{(Xinyu Huang and Tianhao Cheng contributed equally. Corresponding author: Xiaobo Zhang.)}}
\thanks{Weiwei Tian, Rui Feng, and Daoying Geng are with the Academy for Engineering and Technology, Fudan University, No. 220 Handan Road, Shanghai 200433, China~(e-mail: \{wwtian20, fengrui\}@fudan.edu.cn, gengdy@163.com).}
\thanks{Xinyu Huang, Tianhao Cheng, and Rui Feng are with the School of Computer Science, Shanghai Key Laboratory of Intelligent Information Processing, Fudan University, No. 2005 Songhu Road, Shanghai 200438, China~(e-mail: xinyuhuang20@fudan.edu.cn, thcheng23@m.fudan.edu.cn).}
\thanks{Wen He, Rui Feng, and Xiaobo Zhang are with the Department of Respiratory Medicine, Children’s Hospital of Fudan University, No. 399 Wanyuan Road, Shanghai 201102, China~(e-mail: hewen@fudan.edu.cn, zhangxiaobo0307@163.com).}
\thanks{Jinwu Fang is with the School of Public Health, Fudan University, No. 130 Dongan Road, Shanghai 200032, China~(e-mail: fangjinwu007@126.com).}
\thanks{Daoying Geng is also with the Department of Radiology, Huashan Hospital, Fudan University, No. 12 Wulumuqi Rd. Middle, Shanghai 200040, China.}
}

\maketitle

\begin{abstract}
Pediatric pneumonia is the leading cause of death among children under five years worldwide, imposing a substantial burden on affected families. Currently, there are three significant hurdles in diagnosing and treating pediatric pneumonia. Firstly, pediatric pneumonia shares similar symptoms with other respiratory diseases, making rapid and accurate differential diagnosis challenging. Secondly, primary hospitals often lack sufficient medical resources and experienced doctors. Lastly, providing personalized diagnostic reports and treatment recommendations is labor-intensive and time-consuming. To tackle these challenges, we proposed a \textbf{Med}ical \textbf{M}ultimodal \textbf{L}arge \textbf{L}anguage \textbf{M}odel for \textbf{P}ediatric \textbf{P}neumonia~(P2Med-MLLM). This was the first foundation model tailored for patients primarily diagnosed with pediatric pneumonia, capable of handling diverse clinical tasks—such as generating free-text radiology reports and medical records—within a unified framework. Specifically, P2Med-MLLM can process both pure text and image-text data, trained on an extensive and large-scale dataset~(P2Med-MD), including real clinical information from 163,999 outpatient and 8,684 inpatient cases. This dataset comprised 2D chest X-ray images, 3D chest Computed Tomography~(CT) images, corresponding radiology reports, and outpatient and inpatient records. P2Med-MLLM combined a Large Language Model~(LLM) with a vision encoder, fine-tuning them together to handle multiple temporally sequenced and interleaved 2D or 3D images with corresponding radiology reports using a perceiver module. We designed a three-stage training strategy to enable P2Med-MLLM to comprehend medical knowledge and follow instructions for various clinical tasks. To rigorously evaluate P2Med-MLLM’s performance, we developed P2Med-MBench, a benchmark consisting of 642 meticulously verified samples by pediatric pulmonology specialists, covering six clinical decision-support tasks and a balanced variety of diseases. The automated scoring results demonstrated the superiority of P2Med-MLLM. This work plays a crucial role in assisting primary care doctors with prompt disease diagnosis and treatment planning, alleviating patient disease burden, reducing severe symptom mortality rates, and optimizing the allocation of medical resources.
\end{abstract}
\section{Introduction}

\begin{figure*}[htbp]
  \centering
  \vspace{-0.9em}
  \begin{subfigure}[b]{0.49\textwidth}
    \begin{tikzpicture}
      \node[anchor=south west,inner sep=0] (image) at (0,0) {\includegraphics[height=10.0cm, width=\linewidth]{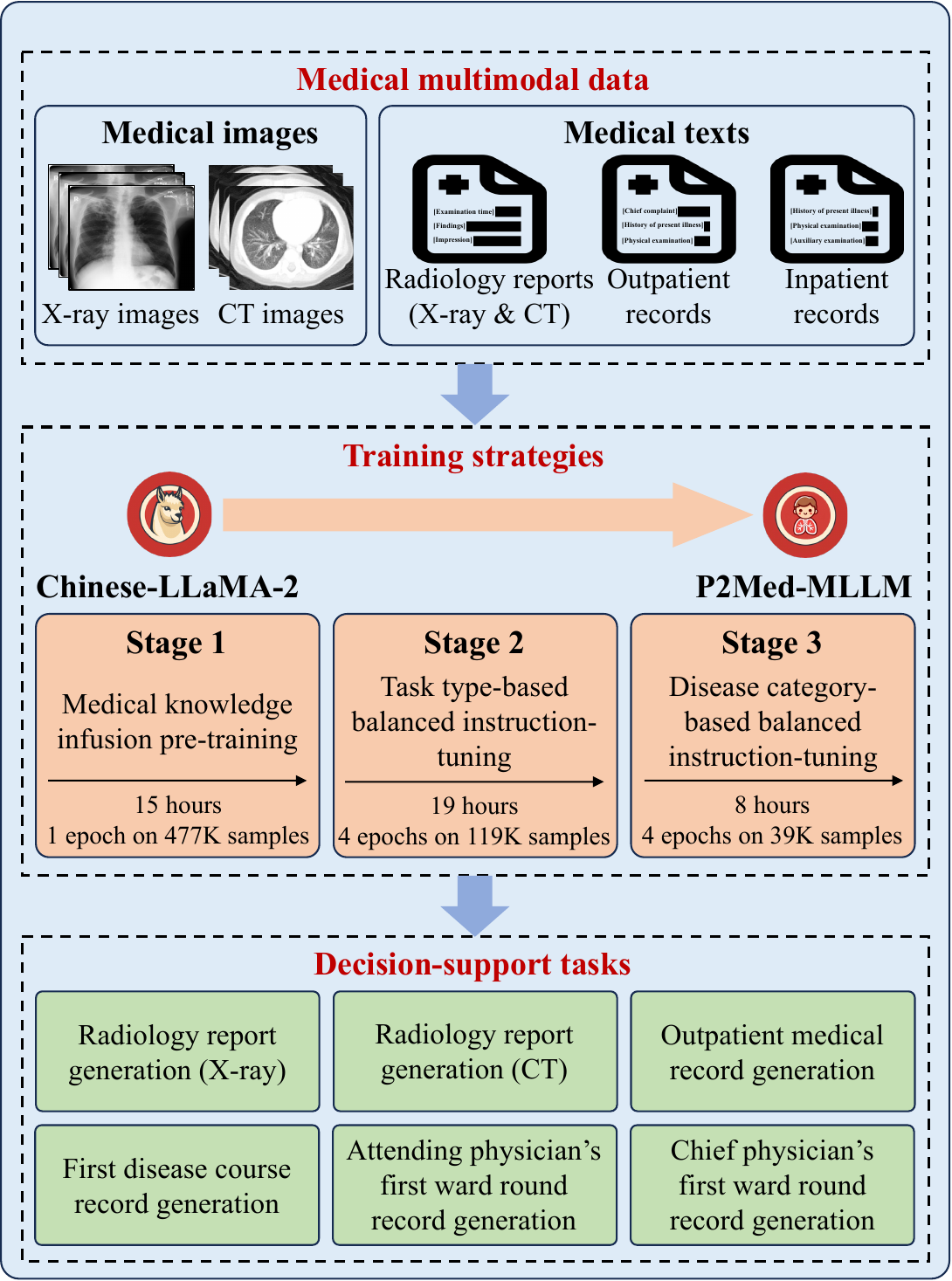}};
      \node[anchor=north west] at (image.north west) {\textbf{a}}; 
    \end{tikzpicture}
    \caption{}
    \label{overview_flowchart}
  \end{subfigure}%
  \hfill
  \begin{subfigure}[b]{0.49\textwidth}
    \begin{tikzpicture}
      \node[anchor=south west,inner sep=0] (image) at (0,0) {\includegraphics[height=10.0cm, width=\linewidth]{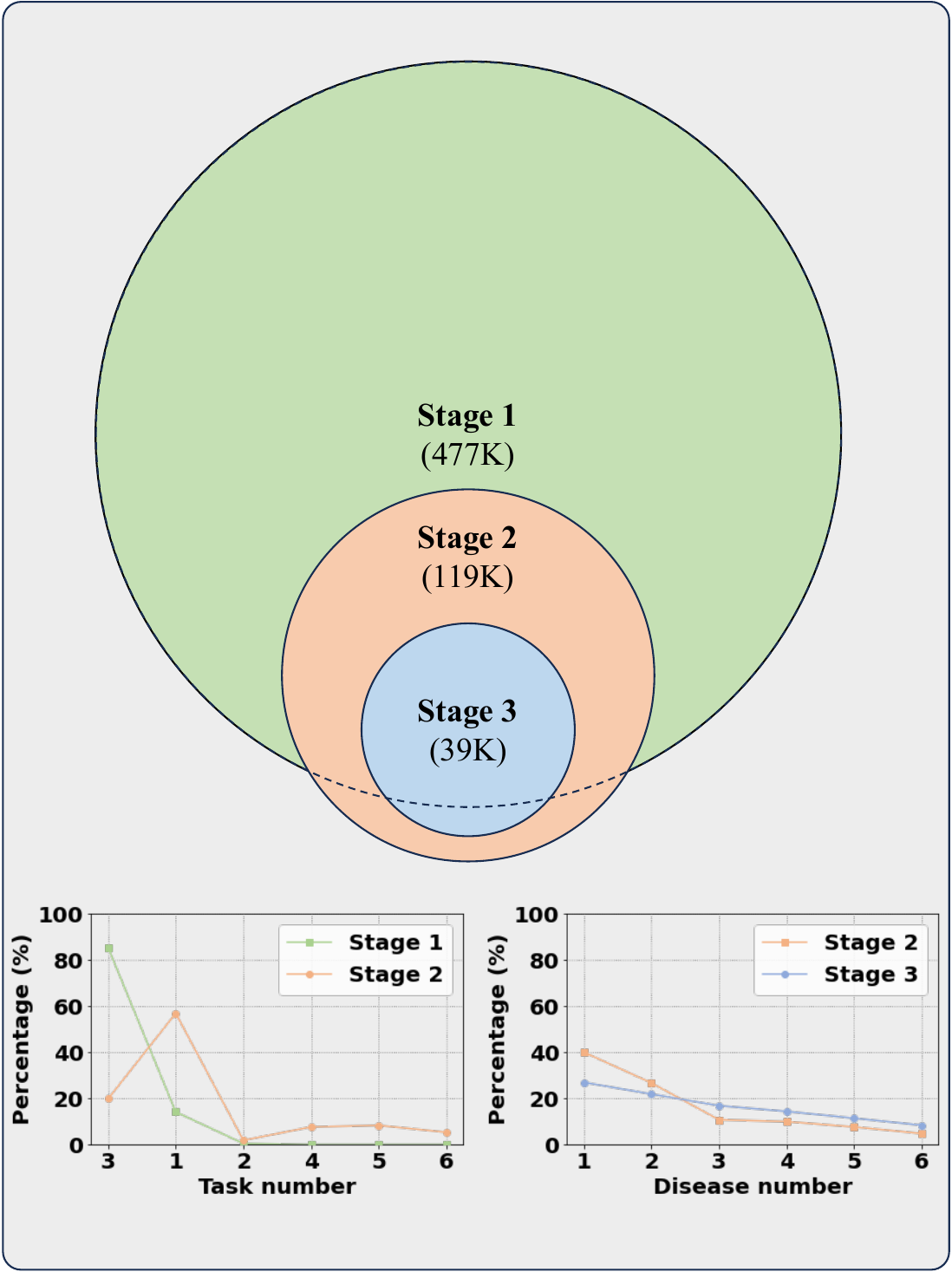}};
      \node[anchor=north west] at (image.north west) {\textbf{b}}; 
    \end{tikzpicture}
    \caption{}
    \label{overview_dataset}
  \end{subfigure}
  \vspace{-2em}
  
  \caption{\textbf{Overview.} (\textbf{a}) The flowchart of this study. (\textbf{b}) The data distribution in different training stages. Note: P2Med-MLLM: Medical Multimodal Large Language Model for Pediatric Pneumonia. CT: Computed Tomography.}
  \label{overview_fig}
\end{figure*}

In 2021 alone, more than 0.5 million children under the age of five died from Lower Respiratory Infections~(LRI) worldwide, accounting for 12\% of total deaths~\cite{roser2021child}. Among LRI, pediatric pneumonia, especially when accompanied by severe symptoms and complications, has the highest morbidity and mortality, particularly in developing countries~\cite{ebeledike2022pediatric}. Pediatric pneumonia, bronchitis, and asthma share similar symptoms like coughing and wheezing, making prompt diagnosis upon admission very challenging~\cite{yu2021identification}. Limited healthcare resources and a lack of experienced doctors in primary hospitals exacerbate this situation, leading to misdiagnoses and inappropriate treatments.

To meet the growing demands of precision medicine, deep learning-based technologies have emerged for identifying pediatric respiratory diseases~\cite{yu2020role, yu2021identification}, early triage~\cite{liang2020early}, and predicting clinical outcomes~\cite{zhou2023transformer}. Despite achieving or nearing human expert levels, these models primarily treated clinical tasks as simple classification or regression problems, falling short of providing detailed and reliable diagnostic bases and treatment recommendations.

Recently, Multimodal Large Language Models~(MLLMs) have experienced exponential growth in general domains~\cite{alayrac2022flamingo, awadalla2023openflamingo, yang2023dawn, wu2023next, liu2024visual, liu2024improved}, but they were still not fully capable of effectively supporting real-world clinical applications~\cite{wu2023can}. The essential reason was that, to protect patient privacy, these models were mainly trained on medical textbooks and literature from the internet, without exposure to real and comprehensive medical data. Aligning with human doctors has significantly improved MLLMs' performance across various specialties~(\textit{e.g.}, radiology, pathology, ophthalmology, and dermatology) and tasks~(\textit{e.g.}, disease diagnosis, medical image generation, medical image caption, medical report generation, medical report summarization, rationale diagnosis, survival prediction, medical image-text retrieval, medical report quality assessment, medical question answering, and medical visual question answering)~\cite{moor2023med, liu2023medical, wu2023towards, liu2023qilin, zhou2023path, zhou2023skingpt, gao2023ophglm, xu2023elixr, leellm, tu2024towards, li2024llava, tu2024generalist, lu2024multimodal, li2024integrated, zhao2024chatcad+, bannur2024maira} in the healthcare field. Inspired by the aforementioned work, we aim to explore the feasibility of MLLMs using real clinical data on pediatric pneumonia. Given the complexity of pediatric pneumonia, we mainly face challenges from three aspects:
\begin{itemize}

\item \textbf{Lack of a large-scale and high-quality multimodal dataset for training:} Due to the rapid physical development of children, their medical imaging, laboratory tests, and demographic data significantly differ from those of adults. Currently, there is a shortage of a large-scale pediatric pneumonia dataset that reflects real clinical scenarios. Moreover, real-world data tend to be noisy and have a long-tailed distribution, which can severely impact model training effectiveness.

\item \textbf{Lack of a unified and compatible model architecture:} Addressing the diverse clinical needs in the full process of diagnosing and treating pediatric pneumonia requires a model architecture that can efficiently handle different modalities, sequences, and time-series data inputs, and produce outputs for various tasks in a unified manner. Currently, such a compatible model structure is lacking.

\item \textbf{Lack of a comprehensive and objective evaluation benchmark:} A comprehensive and objective benchmark is crucial for supervising model training and assessing performance. There is a lack of a high-quality evaluation benchmark that covers a wide range of clinical tasks and disease categories.

\end{itemize}

To address the obstacles of applying MLLMs to pediatric pneumonia, we developed a \textbf{Med}ical \textbf{M}ultimodal \textbf{L}arge \textbf{L}anguage \textbf{M}odel tailored for \textbf{P}ediatric \textbf{P}neumonia~(P2Med-MLLM, Fig.~\ref{overview_flowchart}), which was trained and deployed on a local server within the hospital environment to ensure data security and privacy.

To effectively train P2Med-MLLM, we constructed the first large-scale Chinese \textbf{Med}ical \textbf{M}ultimodal \textbf{D}ataset for \textbf{P}ediatric \textbf{P}neumonia~(P2Med-MD), covering real clinical information from 163,999 outpatients and 8,684 inpatients. Specifically, we collected comprehensive medical data for patients with a primary diagnosis of pediatric pneumonia, including 2D chest X-ray and 3D chest Computed Tomography~(CT) images, corresponding radiology reports, outpatient records, and three-level inpatient records reflecting disease progression~\cite{ying2021comparative}~(Fig.~\ref{overview_flowchart}). During the three training stages, we ensured data quality by deduplication, task type-based balanced sampling, and disease category-based balanced sampling~(Fig.~\ref{overview_dataset}).

As for the architecture of P2Med-MLLM, it consisted of three core components: a Large Language Model~(LLM, Chinese-LLaMA-2~\cite{cui2023efficient}), a CLIP-pretrained vision encoder~\cite{radford2021learning}, and a perceiver module~\cite{alayrac2022flamingo}. This design enabled P2Med-MLLM to retain its original capabilities in understanding and generating pure text (outpatient and inpatient records), while also allowing it to interleave multiple 2D chest X-rays or 3D chest CT images with corresponding radiology reports. This facilitated comparative analysis of a patient's radiological examinations over different time points, aligning more closely with clinical practice.

For evaluation, we initialized a \textbf{Med}ical \textbf{M}ultimodal \textbf{Bench}mark for \textbf{P}ediatric \textbf{P}neumonia, termed P2Med-MBench. This benchmark covered various disease categories and valuable clinical tasks, including radiology report generation~(X-ray), radiology report generation~(CT), outpatient medical record generation, first disease course record generation, attending physician's first ward round record generation, and chief physician's first ward round record generation~(Fig.~\ref{overview_flowchart}). All real-world data have been meticulously verified by professional pediatric pulmonology specialists to ensure quality and representativeness. Automatic scoring of results generated by P2Med-MLLM and other open-source LLMs on P2Med-MBench, along with a series of ablation studies, demonstrated the superiority of our approach.

Overall, the main contributions of our work are summarized as follows:
\begin{itemize}

\item \textbf{Construct a large-scale and high-quality multimodal dataset~(P2Med-MD):} Based on real clinical scenarios, we developed the first Chinese medical multimodal dataset for patients with a primary diagnosis of pediatric pneumonia. This dataset covered various diseases and clinical tasks. 

\item \textbf{Propose a unified and compatible multimodal model architecture~(P2Med-MLLM):} For different clinical tasks, we introduced the first model capable of handling both pure text data~(outpatient and inpatient records) and temporally sequenced, interleaved image-text pairs~(2D X-rays and 3D CT images, along with corresponding radiology reports).

\item \textbf{Establish a comprehensive and objective multimodal evaluation benchmark~(P2Med-MBench):} To supervise the training process and objectively evaluate different models, we meticulously designed a multimodal benchmark with balanced distributions of diseases and clinical tasks. Extensive quantitative and qualitative experimental results demonstrated the effectiveness of our approach.

\end{itemize}
\section{Results}

In this section, we conducted experiments on various tasks, including radiology report generation~(X-ray), radiology report generation~(Computed Tomography, CT), outpatient medical record generation, first disease course record generation, attending physician's first ward round record generation, and chief physician's first ward round record generation. We began by describing the evaluation metrics used for the experiments. Then, we presented the quantitative and qualitative results of our framework on the Medical Multimodal Benchmark for Pediatric Pneumonia~(P2Med-MBench).

\subsection{Evaluation Metrics}

\begin{figure*}[!t]
\centerline{\includegraphics[width=\textwidth]{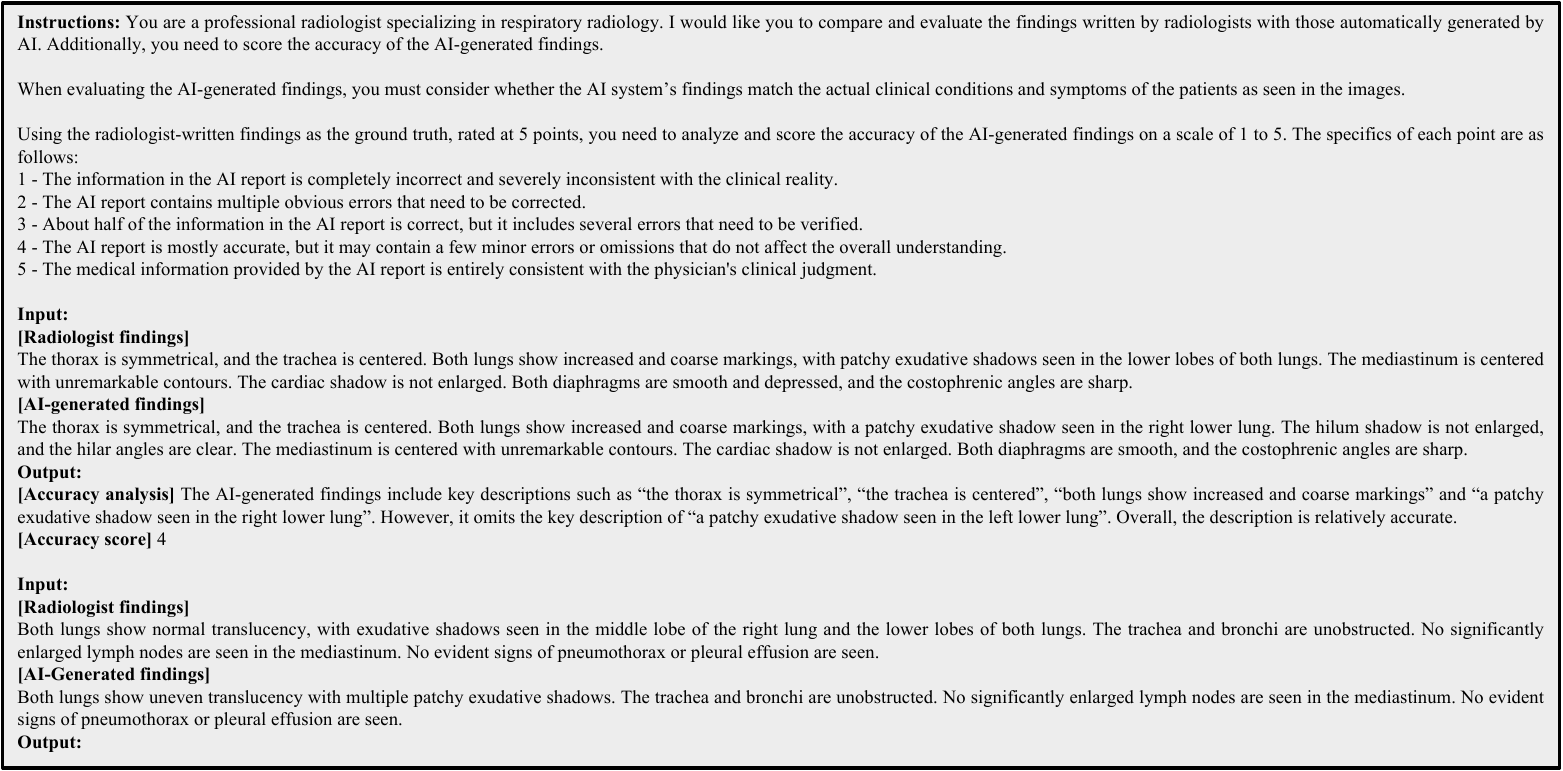}}
\caption{\textbf{Illustration of the evaluation process~(English version).} We evaluated model-generated answers using 13B Chinese-LLaMA-2.}
\label{automatic_rating_english}
\end{figure*}

To evaluate the professional performance of various Medical Multimodal Large Language Model variants for Pediatric Pneumonia~(P2Med-MLLM) and baseline models, we utilized 13B Chinese-LLaMA-2~\cite{cui2023efficient} with a one-shot in-context example to automatically score the generated open-ended responses and provide reasons for the given scores. Our pediatric pulmonology specialists meticulously curated a set of examples across tasks and evaluation components based on their clinical expertise. For the most critical evaluation components in a range of tasks, such as impression or diagnosis results, we employed accuracy and comprehensiveness metrics. For other evaluation components, we used accuracy alone. We assessed the quality of the generated answers using a 5-point scale, as depicted in Fig.~\ref{automatic_rating_english}. The original Chinese version can be found in Fig.~\ref{automatic_rating_chinese}. The 95\% confidence interval for each metric was calculated using the $t$-distribution.

\begin{table*}[h]
\begin{threeparttable}
\caption{\textbf{Comparison with baseline models in six tasks.} Accuracy and Comprehensiveness scores of impression or diagnosis results were reported, representing the key metrics of evaluation. The metrics presented reflected the average scores across all test samples, with 95\% confidence intervals in parentheses. The best results were bolded. Task 1 to task 6 represented radiology report generation~(X-ray), radiology report generation~(CT), outpatient medical record generation, first disease course record generation, attending physician's first ward round record generation, and chief physician's first ward round record generation, respectively. The baseline models were two recent Chinese LLMs~(Baichuan 2 and Chinese-LLaMA-2).}
\label{SOTA}
\renewcommand\arraystretch{1.5}
\setlength{\tabcolsep}{3.4pt}
\tiny
\begin{tabular}{lllllllllllllllll}
\hline
\multirow{2}{*}{Method}          & \multirow{2}{*}{Size} & \multirow{2}{*}{Year} & \multicolumn{2}{l}{Task 1}                                                                                                                  & \multicolumn{2}{l}{Task 2}                                                                                                                  & \multicolumn{2}{l}{Task 3}                                                                                                                  & \multicolumn{2}{l}{Task 4}                                                                                                                  & \multicolumn{2}{l}{Task 5}                                                                                                                  & \multicolumn{2}{l}{Task 6}                                                                                                                  & \multicolumn{2}{l}{Average}                                                                                                                     \\ \cmidrule(r){4-5} \cmidrule(r){6-7} \cmidrule(r){8-9} \cmidrule(r){10-11} \cmidrule(r){12-13} \cmidrule(r){14-15} \cmidrule(r){16-17} 
                                 &                       &                       & Acc                                                                  & Comp                                                                 & Acc                                                                  & Comp                                                                 & Acc                                                                  & Comp                                                                 & Acc                                                                  & Comp                                                                 & Acc                                                                  & Comp                                                                 & Acc                                                                  & Comp                                                                 & Acc                                                                  & Comp                                                                 \\ \hline
\multirow{3}{*}{Baichuan 2~\cite{yang2023baichuan}}      & 7B                    & 2023                  & -                                                                    & -                                                                    & -                                                                    & -                                                                    & \begin{tabular}[c]{@{}l@{}}2.29\\ (2.07, 2.51)\end{tabular}          & \begin{tabular}[c]{@{}l@{}}2.89\\ (2.64, 3.14)\end{tabular}          & \begin{tabular}[c]{@{}l@{}}3.42\\ (3.18, 3.66)\end{tabular}          & \begin{tabular}[c]{@{}l@{}}3.64\\ (3.40, 3.88)\end{tabular}          & \begin{tabular}[c]{@{}l@{}}2.83\\ (2.59, 3.07)\end{tabular}          & \begin{tabular}[c]{@{}l@{}}3.00\\ (2.72, 3.28)\end{tabular}          & \begin{tabular}[c]{@{}l@{}}2.67\\ (2.48, 2.86)\end{tabular}          & \begin{tabular}[c]{@{}l@{}}2.98\\ (2.73, 3.23)\end{tabular}          & \begin{tabular}[c]{@{}l@{}}2.80\\ (2.68, 2.92)\end{tabular}          & \begin{tabular}[c]{@{}l@{}}3.13\\ (3.00, 3.26)\end{tabular}          \\ \cline{2-17} 
                                 & 13B                   & 2023                  & -                                                                    & -                                                                    & -                                                                    & -                                                                    & \begin{tabular}[c]{@{}l@{}}2.96\\ (2.76, 3.16)\end{tabular}          & \begin{tabular}[c]{@{}l@{}}3.26\\ (3.01, 3.51)\end{tabular}          & \begin{tabular}[c]{@{}l@{}}3.48\\ (3.26, 3.70)\end{tabular}          & \begin{tabular}[c]{@{}l@{}}3.55\\ (3.31, 3.79)\end{tabular}          & \begin{tabular}[c]{@{}l@{}}3.00\\ (2.76, 3.24)\end{tabular}          & \begin{tabular}[c]{@{}l@{}}3.43\\ (3.16, 3.70)\end{tabular}          & \textbf{\begin{tabular}[c]{@{}l@{}}2.96\\ (2.75, 3.17)\end{tabular}} & \begin{tabular}[c]{@{}l@{}}3.30\\ (3.03, 3.57)\end{tabular}          & \begin{tabular}[c]{@{}l@{}}3.10\\ (2.99, 3.21)\end{tabular}          & \begin{tabular}[c]{@{}l@{}}3.38\\ (3.25, 3.51)\end{tabular}          \\ \hline\multirow{3}{*}{Chinese-LLaMA-2~\cite{cui2023efficient}} & 7B                    & 2023                  & -                                                                    & -                                                                    & -                                                                    & -                                                                    & \begin{tabular}[c]{@{}l@{}}1.50\\ (1.17, 1.83)\end{tabular}          & \begin{tabular}[c]{@{}l@{}}1.32\\ (1.03, 1.61)\end{tabular}          & \begin{tabular}[c]{@{}l@{}}2.07\\ (1.69, 2.45)\end{tabular}          & \begin{tabular}[c]{@{}l@{}}2.06\\ (1.68, 2.44)\end{tabular}          & \begin{tabular}[c]{@{}l@{}}2.69\\ (2.44, 2.94)\end{tabular}          & \begin{tabular}[c]{@{}l@{}}2.56\\ (2.33, 2.79)\end{tabular}          & \begin{tabular}[c]{@{}l@{}}2.54\\ (2.29, 2.79)\end{tabular}          & \begin{tabular}[c]{@{}l@{}}2.61\\ (2.36, 2.86)\end{tabular}          & \begin{tabular}[c]{@{}l@{}}2.20\\ (2.04, 2.36)\end{tabular}          & \begin{tabular}[c]{@{}l@{}}2.14\\ (1.99, 2.29)\end{tabular}          \\ \cline{2-17} 
                                 & 13B                   & 2023                  & -                                                                    & -                                                                    & -                                                                    & -                                                                    & \begin{tabular}[c]{@{}l@{}}2.68\\ (2.47, 2.89)\end{tabular}          & \begin{tabular}[c]{@{}l@{}}2.91\\ (2.69, 3.13)\end{tabular}          & \begin{tabular}[c]{@{}l@{}}3.26\\ (2.98, 3.54)\end{tabular}          & \begin{tabular}[c]{@{}l@{}}3.07\\ (2.80, 3.34)\end{tabular}          & \begin{tabular}[c]{@{}l@{}}2.56\\ (2.28, 2.84)\end{tabular}          & \begin{tabular}[c]{@{}l@{}}2.41\\ (2.20, 2.62)\end{tabular}          & \begin{tabular}[c]{@{}l@{}}2.18\\ (1.97, 2.39)\end{tabular}          & \begin{tabular}[c]{@{}l@{}}2.32\\ (2.11, 2.53)\end{tabular}          & \begin{tabular}[c]{@{}l@{}}2.67\\ (2.54, 2.80)\end{tabular}          & \begin{tabular}[c]{@{}l@{}}2.68\\ (2.56, 2.80)\end{tabular}          \\ \hline
P2Med-MLLM                       & 8B                    & Ours                  & \textbf{\begin{tabular}[c]{@{}l@{}}3.04\\ (2.75, 3.33)\end{tabular}} & \textbf{\begin{tabular}[c]{@{}l@{}}3.09\\ (2.83, 3.35)\end{tabular}} & \textbf{\begin{tabular}[c]{@{}l@{}}3.81\\ (3.50, 4.12)\end{tabular}} & \textbf{\begin{tabular}[c]{@{}l@{}}3.18\\ (2.97, 3.39)\end{tabular}} & \textbf{\begin{tabular}[c]{@{}l@{}}3.37\\ (3.20, 3.54)\end{tabular}} & \textbf{\begin{tabular}[c]{@{}l@{}}4.17\\ (4.04, 4.30)\end{tabular}} & \textbf{\begin{tabular}[c]{@{}l@{}}3.73\\ (3.47, 3.99)\end{tabular}} & \textbf{\begin{tabular}[c]{@{}l@{}}3.99\\ (3.72, 4.26)\end{tabular}} & \textbf{\begin{tabular}[c]{@{}l@{}}3.07\\ (2.80, 3.34)\end{tabular}} & \textbf{\begin{tabular}[c]{@{}l@{}}3.55\\ (3.24, 3.86)\end{tabular}} & \begin{tabular}[c]{@{}l@{}}2.88\\ (2.61, 3.15)\end{tabular}          & \textbf{\begin{tabular}[c]{@{}l@{}}3.47\\ (3.15, 3.79)\end{tabular}} & \textbf{\begin{tabular}[c]{@{}l@{}}3.32\\ (3.21, 3.43)\end{tabular}} & \textbf{\begin{tabular}[c]{@{}l@{}}3.57\\ (3.46, 3.68)\end{tabular}} \\ \hline
\end{tabular}
\begin{tablenotes}[para,flushleft]
\item Note: CT: Computed Tomography. LLMs: Large Language Models. Acc: Accuracy. Comp: Comprehensiveness. P2Med-MLLM: Medical Multimodal Large Language Model for Pediatric Pneumonia.
\end{tablenotes}
\end{threeparttable}
\end{table*}

\begin{figure*}[!t]
\centerline{\includegraphics[width=\textwidth]{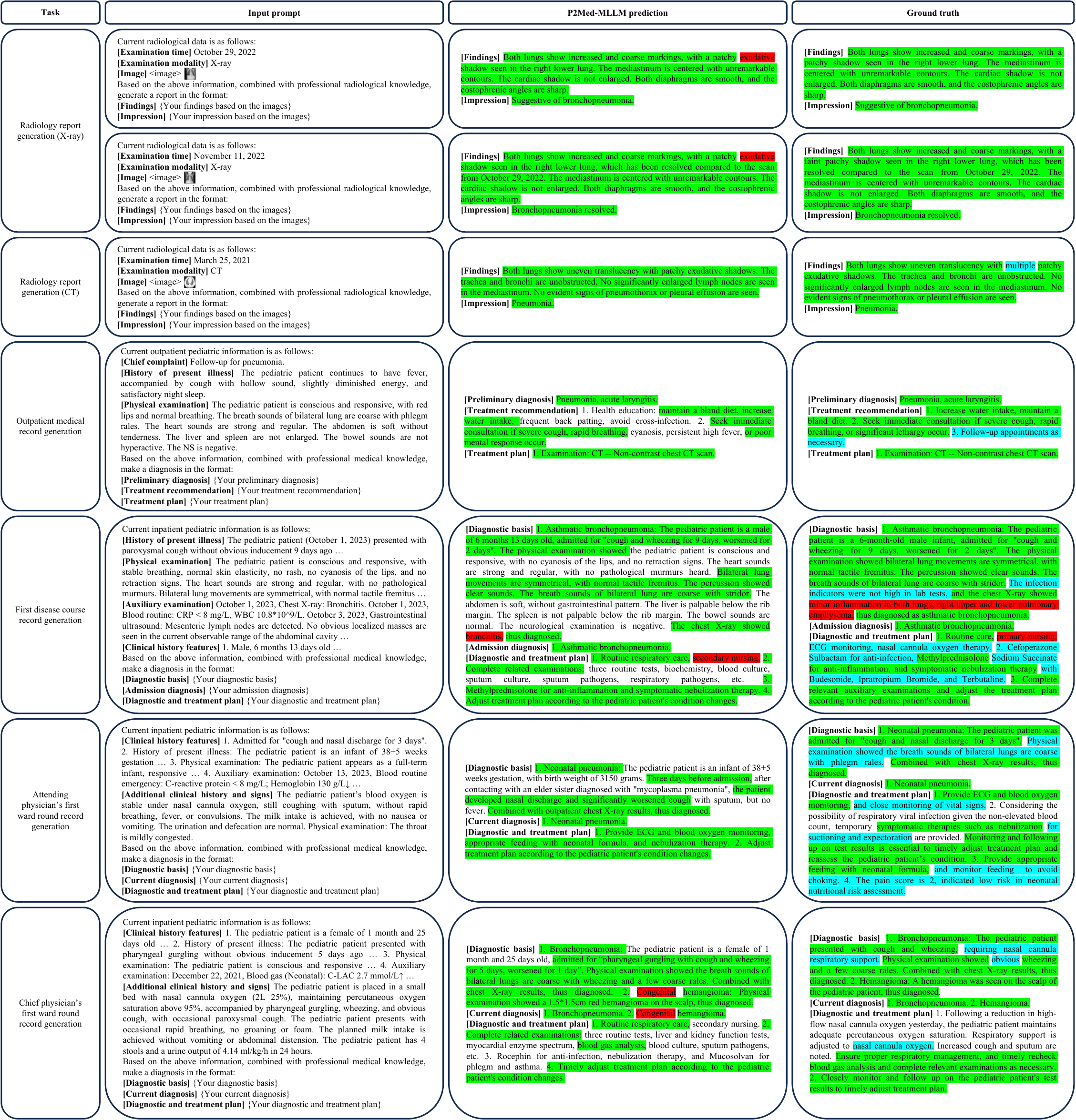}}
\caption{\textbf{Qualitative examples of six different evaluation tasks~(English version).} We presented input prompts along with answers generated by P2Med-MLLM and the target ground truth. The green color in the figure highlighted correct predictions, the red color indicated errors, and the blue color denoted neglected parts. Note: P2Med-MLLM: Medical Multimodal Large Language Model for Pediatric Pneumonia. CT: Computed Tomography.}
\label{qualitative_results_english}
\end{figure*}

\subsection{Radiology Report Generation~(X-ray)}
As shown in Table \ref{SOTA}, Large Language Models~(LLMs) such as Baichuan 2~\cite{yang2023baichuan} and Chinese-LLaMA-2~\cite{cui2023efficient} can only process pure texts. In clinical practice, X-ray images were crucial for screening and diagnosing pediatric pneumonia. By incorporating a perceiver with an LLM, P2Med-MLLM can handle sequential 2D X-ray images and generate corresponding radiology reports. Fig.~\ref{qualitative_results_english} demonstrated that P2Med-MLLM was capable of processing X-ray images taken at different times from the same patient in two conversation turns. The model generated different impressions: ``\textit{bronchopneumonia}"~(October 29, 2022) and ``\textit{bronchopneumonia resolved}"~(November 11, 2022), effectively reflecting the patient's disease progression. The original Chinese results were detailed in Fig.~\ref{qualitative_results_chinese}.

\subsection{Radiology Report Generation~(CT)}
In addition to 2D X-ray images, P2Med-MLLM can also generate radiology reports for 3D CT images. As shown in Fig.~\ref{qualitative_results_english} and Fig.~\ref{qualitative_results_chinese}, the model successfully identified critical radiological features in the images and recognized underlying diseases.

\subsection{Outpatient Medical Record Generation}
Generating outpatient medical records was a challenging and open-ended task that required comprehensive analysis of the outpatient's chief complaint, history of present illness, and physical examination. As indicated in Table \ref{SOTA}, the 8B P2Med-MLLM demonstrated significant improvements compared to other LLMs~(such as the 7B or 13B Baichuan 2 or Chinese-LLaMA-2). For example, P2Med-MLLM increased the accuracy of preliminary diagnosis from 2.96 to 3.37 and improved the comprehensiveness from 3.26 to 4.17. These results indicated that P2Med-MLLM can generate radiology reports without compromising the language model's capabilities. Fig.~\ref{qualitative_results_english} and Fig.~\ref{qualitative_results_chinese} showed that P2Med-MLLM can make accurate diagnoses in free-text format and provide highly relevant treatment recommendations and plans, despite missing information ``\textit{follow-up appointments as necessary}".

\subsection{First Disease Course Record Generation}
Generating resident physician's first disease course records for inpatients was more challenging because it required comprehensive analysis of various information, including the history of present illness, physical examination, auxiliary examinations, and clinical history features. As depicted in Table \ref{SOTA}, compared to the suboptimal models, the 7B or 13B Baichuan 2, P2Med-MLLM showed an improvement in the accuracy and comprehensiveness of admission diagnosis by 0.25 and 0.35, respectively. Qualitatively, as shown in Fig.~\ref{qualitative_results_english} and Fig.~\ref{qualitative_results_chinese}, P2Med-MLLM can understand the provided patient information and questions, generating a relatively accurate diagnostic basis, admission diagnosis, and diagnostic and treatment plan in a standardized format. However, some details still needed improvement. For instance, the chest X-ray on October 1, 2023, showed ``\textit{bronchitis}", while the chest X-ray on October 8, 2023, showed ``\textit{minor inflammation in both lungs and right upper and lower pulmonary emphysema}". The most recent examination results should be prioritized. Additionally, there were some errors in the nursing level recommendations and medication guidelines that needed to be addressed.

\subsection{Attending Physician's First Ward Round Record Generation}
Generating attending physician's first ward round records for inpatients was also a crucial and meaningful task for generative medical foundation models. This task involved using input clinical history features and additional clinical history and signs to produce the patient's diagnostic basis, current diagnosis, and diagnostic and treatment plan. In Table \ref{SOTA}, the 8B P2Med-MLLM outperformed the 13B Baichuan 2 by 0.07 in accuracy and 0.12 in comprehensiveness for the current diagnosis, highlighting the advantages of our model. Fig.~\ref{qualitative_results_english} and Fig.~\ref{qualitative_results_chinese} demonstrated the effectiveness of P2Med-MLLM in generating the three components for this task. However, the generated record still had some shortcomings, such as omitting critical information like abnormal findings in the ``\textit{physical examination}" section of the diagnostic basis and the ``\textit{neonatal nutritional risk assessment}" in the diagnostic and treatment plan.

\subsection{Chief Physician's First Ward Round Record Generation}
The task of generating chief physician's first ward round records for inpatients was similar to that of generating attending physician's first ward round records. As shown in Table \ref{SOTA}, although the 8B P2Med-MLLM maintained advantages in most tasks, components, and metrics, it fell behind the 13B Baichuan 2 by 0.08 in accuracy for the current diagnosis in this specific task. This was remarkable especially considering that, for the baseline LLMs, namely Baichuan 2 and Chinese-LLaMA-2, the 13B models significantly outperformed the 7B models. It not only demonstrated that more model parameters can lead to further performance improvements, but also indicated the potential of LLMs that can be further simulated by scaling up the models. Due to computational resource constraints, we opted for the largest model size of 8B to balance performance. Fig.~\ref{qualitative_results_english} and Fig.~\ref{qualitative_results_chinese} showed that our model provided correct answers in most components, except for the incorrect addition of ``\textit{congenital}" in the diagnosis of hemangioma and the omission of the ``\textit{nasal cannula oxygen}" keyword in the components of the diagnostic basis and diagnostic and treatment plan.

\section{Discussion}

\subsection{Impact of Different Stages and Modalities in P2Med-MLLM}

\begin{table*}[h]
\begin{threeparttable}
\caption{\textbf{An ablation study of P2Med-MLLM by removing single stage or modality.} Accuracy and Comprehensiveness scores of impression or diagnosis results were reported, representing the key metrics of evaluation. The metrics presented reflected the average scores across all test samples, with 95\% confidence intervals in parentheses. The best results were bolded. Stage 1 to stage 3 represented medical knowledge infusion pre-training, task type-based balanced instruction-tuning, and disease category-based balanced instruction-tuning, respectively.}
\label{ablation_study_stage_modality_main}
\renewcommand\arraystretch{1.5}
\setlength{\tabcolsep}{7.2pt}
\begin{tabular}{llllllll}
\hline
Task description                       & Metric & \begin{tabular}[c]{@{}l@{}}Full \\ P2Med-MLLM\end{tabular}                                             & \begin{tabular}[c]{@{}l@{}} A \\ W/o stage 1\end{tabular}                                                 & \begin{tabular}[c]{@{}l@{}} B \\ W/o stage 2\end{tabular}                                                 & \begin{tabular}[c]{@{}l@{}} C \\ W/o stage 3\end{tabular}                                                 & \begin{tabular}[c]{@{}l@{}} D \\ W/o plain text\end{tabular}                                               & \begin{tabular}[c]{@{}l@{}} E \\ W/o image-text\end{tabular}                                              \\ \hline
\multirow{3}{*}{\begin{tabular}[c]{@{}l@{}}Radiology report\\ generation (X-ray)\end{tabular}}                           & Accuracy    & \textbf{\begin{tabular}[c]{@{}l@{}}3.04\\ (2.75, 3.33)\end{tabular}} & \begin{tabular}[c]{@{}l@{}}3.00\\ (2.74, 3.26)\end{tabular} & \begin{tabular}[c]{@{}l@{}}2.75\\ (2.48, 3.02)\end{tabular} & \begin{tabular}[c]{@{}l@{}}2.72\\ (2.45, 2.99)\end{tabular} & \begin{tabular}[c]{@{}l@{}}2.72\\ (2.45, 2.99)\end{tabular} & -                                                           \\ \cline{2-8} 
 & Comprehensiveness   & \begin{tabular}[c]{@{}l@{}}3.09\\ (2.83, 3.35)\end{tabular} & \textbf{\begin{tabular}[c]{@{}l@{}}3.12\\ (2.87, 3.37)\end{tabular}} & \begin{tabular}[c]{@{}l@{}}2.71\\ (2.45, 2.97)\end{tabular} & \begin{tabular}[c]{@{}l@{}}2.69\\ (2.42, 2.96)\end{tabular} & \begin{tabular}[c]{@{}l@{}}2.62\\ (2.35, 2.89)\end{tabular} & -                                                           \\ \hline
\multirow{3}{*}{\begin{tabular}[c]{@{}l@{}}Radiology report \\ generation~(CT)\end{tabular}}                           & Accuracy    & \begin{tabular}[c]{@{}l@{}}3.81\\ (3.50, 4.12)\end{tabular} & \begin{tabular}[c]{@{}l@{}}2.73\\ (2.45, 3.01)\end{tabular} & \textbf{\begin{tabular}[c]{@{}l@{}}3.97\\ (3.70, 4.24)\end{tabular}} & \begin{tabular}[c]{@{}l@{}}3.15\\ (2.92, 3.38)\end{tabular} & \begin{tabular}[c]{@{}l@{}}2.69\\ (2.38, 3.00)\end{tabular} & -                                                           \\ \cline{2-8} 
   & Comprehensiveness   & \begin{tabular}[c]{@{}l@{}}3.18\\ (2.97, 3.39)\end{tabular} & \begin{tabular}[c]{@{}l@{}}2.79\\ (2.50, 3.08)\end{tabular} & \textbf{\begin{tabular}[c]{@{}l@{}}3.46\\ (3.24, 3.68)\end{tabular}} & \begin{tabular}[c]{@{}l@{}}3.40\\ (3.19, 3.61)\end{tabular} & \begin{tabular}[c]{@{}l@{}}2.79\\ (2.48, 3.10)\end{tabular} & -                                                           \\ \hline
\multirow{3}{*}{\begin{tabular}[c]{@{}l@{}}Outpatient medical \\ record generation\end{tabular}}                            & Accuracy    & \begin{tabular}[c]{@{}l@{}}3.37\\ (3.20, 3.54)\end{tabular} & \begin{tabular}[c]{@{}l@{}}3.30\\ (3.04, 3.56)\end{tabular} & \begin{tabular}[c]{@{}l@{}}2.63\\ (2.26, 3.00)\end{tabular} & \textbf{\begin{tabular}[c]{@{}l@{}}3.40\\ (3.21, 3.59)\end{tabular}} & -                                                           & \begin{tabular}[c]{@{}l@{}}1.39\\ (1.00, 1.78)\end{tabular} \\ \cline{2-8} 
  & Comprehensiveness   & \textbf{\begin{tabular}[c]{@{}l@{}}4.17\\ (4.04, 4.30)\end{tabular}} & \begin{tabular}[c]{@{}l@{}}3.39\\ (3.13, 3.65)\end{tabular} & \begin{tabular}[c]{@{}l@{}}2.77\\ (2.40, 3.14)\end{tabular} & \begin{tabular}[c]{@{}l@{}}4.16\\ (4.03, 4.29)\end{tabular} & -                                                           & \begin{tabular}[c]{@{}l@{}}1.47\\ (1.07, 1.87)\end{tabular} \\ \hline
\multirow{3}{*}{\begin{tabular}[c]{@{}l@{}}First disease course \\ record generation\end{tabular}}                           & Accuracy    & \textbf{\begin{tabular}[c]{@{}l@{}}3.73\\ (3.47, 3.99)\end{tabular}} & \begin{tabular}[c]{@{}l@{}}1.22\\ (0.85, 1.59)\end{tabular} & \begin{tabular}[c]{@{}l@{}}2.19\\ (1.80, 2.58)\end{tabular} & \begin{tabular}[c]{@{}l@{}}3.60\\ (3.30, 3.90)\end{tabular} & -                                                           & \begin{tabular}[c]{@{}l@{}}1.62\\ (1.21, 2.03)\end{tabular} \\ \cline{2-8} 
   & Comprehensiveness   & \textbf{\begin{tabular}[c]{@{}l@{}}3.99\\ (3.72, 4.26)\end{tabular}} & \begin{tabular}[c]{@{}l@{}}1.14\\ (0.79, 1.49)\end{tabular} & \begin{tabular}[c]{@{}l@{}}2.12\\ (1.76, 2.48)\end{tabular} & \begin{tabular}[c]{@{}l@{}}3.44\\ (3.14, 3.74)\end{tabular} & -                                                           & \begin{tabular}[c]{@{}l@{}}1.75\\ (1.31, 2.19)\end{tabular} \\ \hline
\multirow{2}{*}{\begin{tabular}[c]{@{}l@{}}Attending physician's \\ first ward round \\ record generation\end{tabular}}                           & Accuracy    & \textbf{\begin{tabular}[c]{@{}l@{}}3.07\\ (2.80, 3.34)\end{tabular}} & \begin{tabular}[c]{@{}l@{}}0.41\\ (0.18, 0.64)\end{tabular} & \begin{tabular}[c]{@{}l@{}}1.75\\ (1.40, 2.10)\end{tabular} & \begin{tabular}[c]{@{}l@{}}3.04\\ (2.76, 3.32)\end{tabular} & -                                                           & \begin{tabular}[c]{@{}l@{}}1.65\\ (1.28, 2.02)\end{tabular} \\ \cline{2-8} 
   & Comprehensiveness   & \textbf{\begin{tabular}[c]{@{}l@{}}3.55\\ (3.24, 3.86)\end{tabular}} & \begin{tabular}[c]{@{}l@{}}0.44\\ (0.20, 0.68)\end{tabular} & \begin{tabular}[c]{@{}l@{}}2.13\\ (1.71, 2.55)\end{tabular} & \begin{tabular}[c]{@{}l@{}}3.21\\ (2.90, 3.52)\end{tabular} & -                                                           & \begin{tabular}[c]{@{}l@{}}1.93\\ (1.51, 2.35)\end{tabular} \\ \hline
\multirow{2}{*}{\begin{tabular}[c]{@{}l@{}}Chief physician's \\ first ward round \\ record generation\end{tabular}}                           & Accuracy    & \textbf{\begin{tabular}[c]{@{}l@{}}2.88\\ (2.61, 3.15)\end{tabular}} & \begin{tabular}[c]{@{}l@{}}0.55\\ (0.31, 0.79)\end{tabular} & \begin{tabular}[c]{@{}l@{}}1.89\\ (1.57, 2.21)\end{tabular} & \begin{tabular}[c]{@{}l@{}}2.71\\ (2.42, 3.00)\end{tabular} & -                                                           & \begin{tabular}[c]{@{}l@{}}2.24\\ (1.87, 2.61)\end{tabular} \\ \cline{2-8} 
   & Comprehensiveness   & \textbf{\begin{tabular}[c]{@{}l@{}}3.47\\ (3.15, 3.79)\end{tabular}} & \begin{tabular}[c]{@{}l@{}}0.64\\ (0.36, 0.92)\end{tabular} & \begin{tabular}[c]{@{}l@{}}2.43\\ (2.03, 2.83)\end{tabular} & \begin{tabular}[c]{@{}l@{}}2.98\\ (2.66, 3.30)\end{tabular} & -                                                           & \begin{tabular}[c]{@{}l@{}}2.62\\ (2.20, 3.04)\end{tabular} \\ \hline
\multirow{3}{*}{Average}                           & Accuracy    & \textbf{\begin{tabular}[c]{@{}l@{}}3.32\\ (3.21, 3.43)\end{tabular}} & \begin{tabular}[c]{@{}l@{}}1.87\\ (1.72, 2.02)\end{tabular} & \begin{tabular}[c]{@{}l@{}}2.53\\ (2.38, 2.68)\end{tabular} & \begin{tabular}[c]{@{}l@{}}3.10\\ (2.99, 3.21)\end{tabular} & \begin{tabular}[c]{@{}l@{}}2.71\\ (2.51, 2.91)\end{tabular} & \begin{tabular}[c]{@{}l@{}}1.72\\ (1.53, 1.91)\end{tabular} \\ \cline{2-8} 
  & Comprehensiveness   & \textbf{\begin{tabular}[c]{@{}l@{}}3.57\\ (3.46, 3.68)\end{tabular}} & \begin{tabular}[c]{@{}l@{}}1.92\\ (1.77, 2.07)\end{tabular} & \begin{tabular}[c]{@{}l@{}}2.60\\ (2.46, 2.74)\end{tabular} & \begin{tabular}[c]{@{}l@{}}3.31\\ (3.20, 3.42)\end{tabular} & \begin{tabular}[c]{@{}l@{}}2.71\\ (2.51, 2.91)\end{tabular} & \begin{tabular}[c]{@{}l@{}}1.94\\ (1.73, 2.15)\end{tabular} \\ \hline
\end{tabular}
\begin{tablenotes}[para,flushleft]
\item Note: P2Med-MLLM: Medical Multimodal Large Language Model for Pediatric Pneumonia. CT: Computed Tomography.
\end{tablenotes}
\end{threeparttable}
\end{table*}

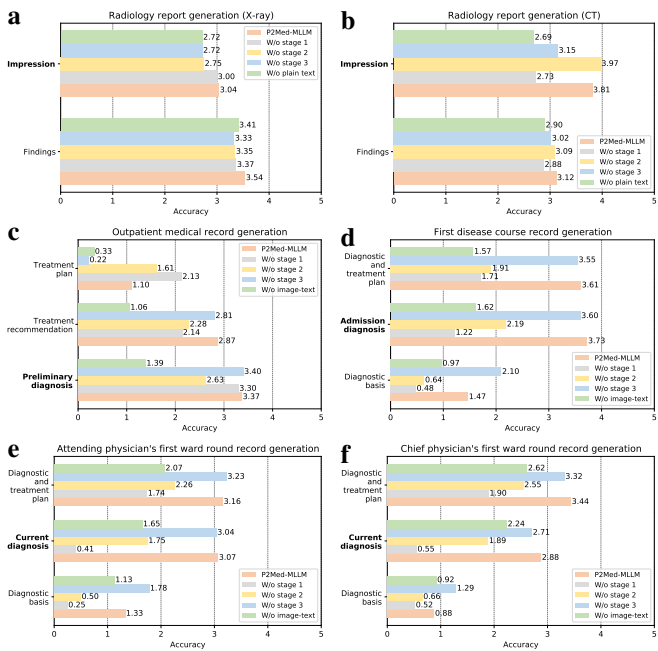
\begin{figure}[htbp]
  \centering
  \begin{subfigure}[b]{0.5\columnwidth}
    \begin{tikzpicture}
      \node[anchor=south west,inner sep=0] (image) at (0,0) {\includegraphics[width=\linewidth]{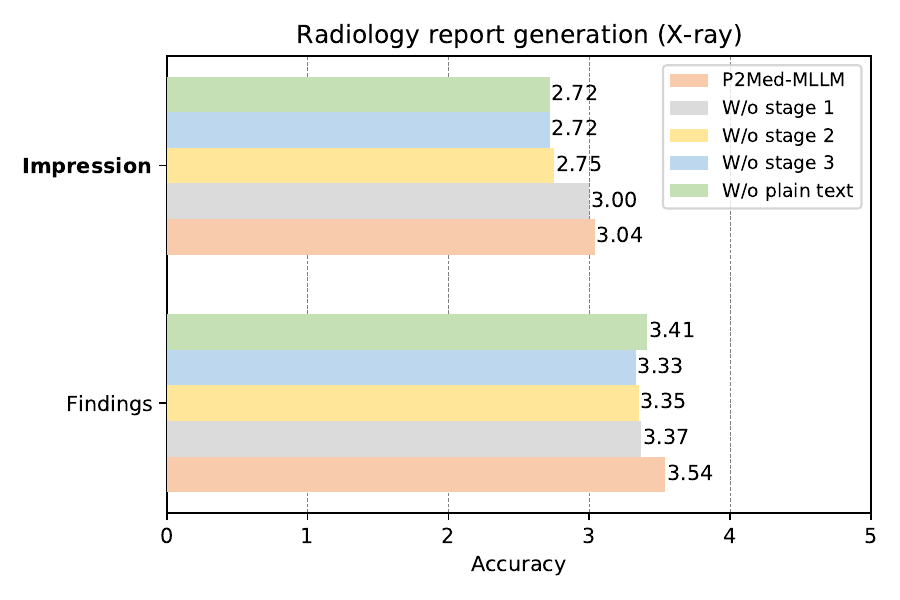}};
      \node[anchor=north west] at (image.north west) {\textbf{a}}; 
    \end{tikzpicture}
    \caption{}
    \label{ablation_stage_modality_task1}
  \end{subfigure}%
  \hfill
  \begin{subfigure}[b]{0.5\columnwidth}
    \begin{tikzpicture}
      \node[anchor=south west,inner sep=0] (image) at (0,0) {\includegraphics[width=\linewidth]{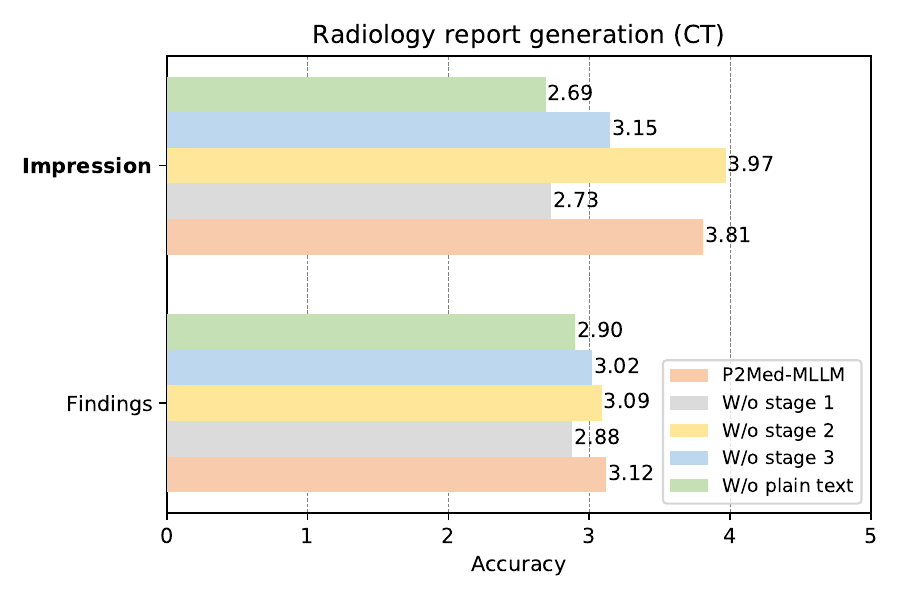}};
      \node[anchor=north west] at (image.north west) {\textbf{b}}; 
    \end{tikzpicture}
    \caption{}
    \label{ablation_stage_modality_task2}
  \end{subfigure}
  \vspace{-3em}

  \begin{subfigure}[b]{0.5\columnwidth}
    \begin{tikzpicture}
      \node[anchor=south west,inner sep=0] (image) at (0,0) {\includegraphics[width=\linewidth]{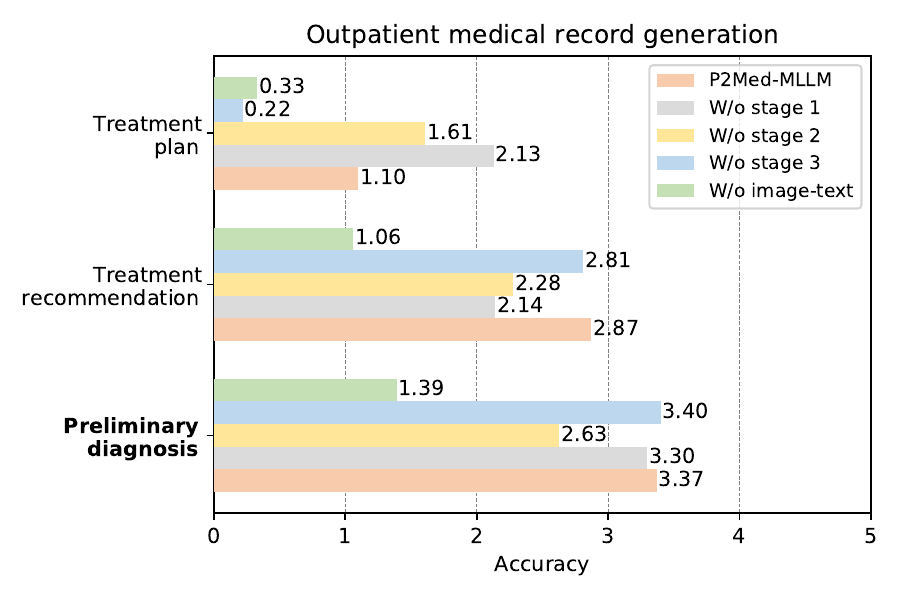}};
      \node[anchor=north west] at (image.north west) {\textbf{c}}; 
    \end{tikzpicture}
    \caption{}
    \label{ablation_stage_modality_task3}
  \end{subfigure}%
  \hfill
  \begin{subfigure}[b]{0.5\columnwidth}
    \begin{tikzpicture}
      \node[anchor=south west,inner sep=0] (image) at (0,0) {\includegraphics[width=\linewidth]{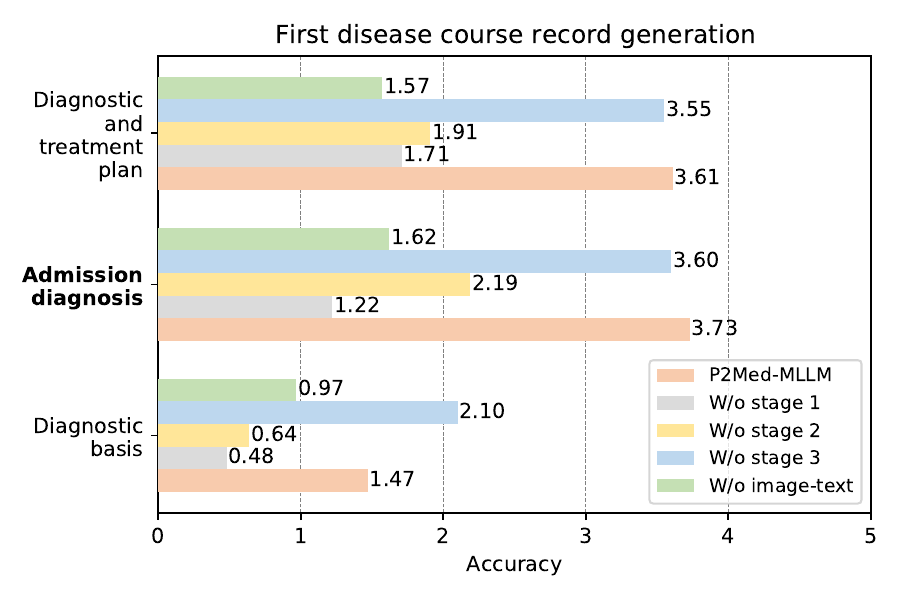}};
      \node[anchor=north west] at (image.north west) {\textbf{d}}; 
    \end{tikzpicture}
    \caption{}
    \label{ablation_stage_modality_task4}
  \end{subfigure}
  \vspace{-3em}

  \begin{subfigure}[b]{0.5\columnwidth}
    \begin{tikzpicture}
      \node[anchor=south west,inner sep=0] (image) at (0,0) {\includegraphics[width=\linewidth]{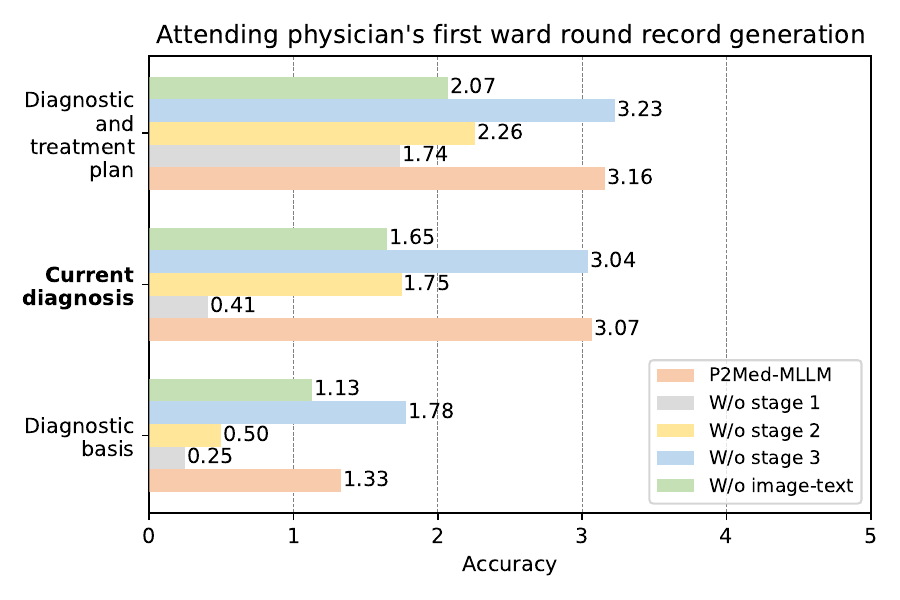}};
      \node[anchor=north west] at (image.north west) {\textbf{e}}; 
    \end{tikzpicture}
    \caption{}
    \label{ablation_stage_modality_task5}
  \end{subfigure}%
  \hfill
  \begin{subfigure}[b]{0.5\columnwidth}
    \begin{tikzpicture}
      \node[anchor=south west,inner sep=0] (image) at (0,0) {\includegraphics[width=\linewidth]{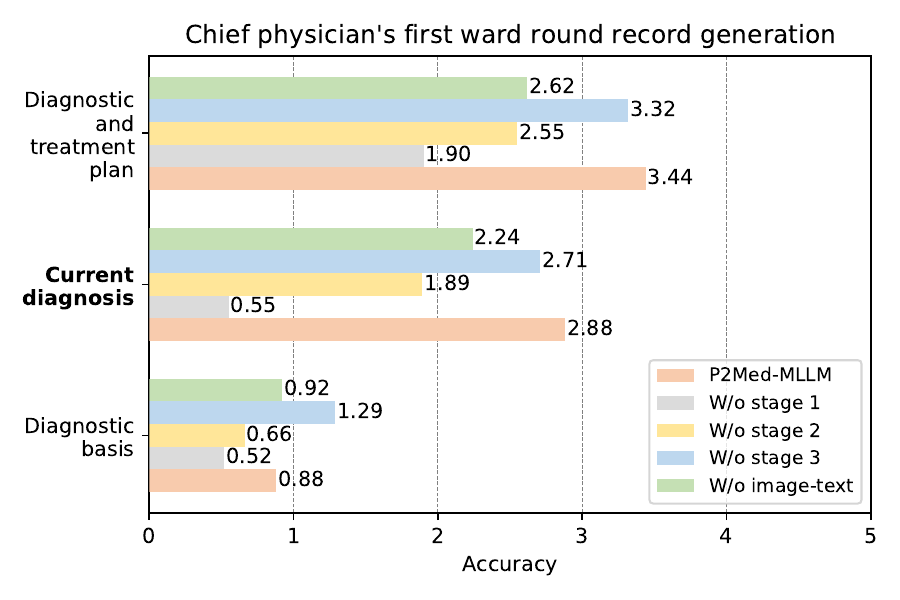}};
      \node[anchor=north west] at (image.north west) {\textbf{f}}; 
    \end{tikzpicture}
    \caption{}
    \label{ablation_stage_modality_task6}
  \end{subfigure}
  \vspace{-3em}
  
  \caption{\textbf{An ablation study of P2Med-MLLM by removing single stage or modality.} We compared six different tasks~(\textbf{a}-\textbf{f}) using the accuracy score, with the most crucial evaluation components highlighted in bold. Stage 1 to stage 3 represented medical knowledge infusion pre-training, task type-based balanced instruction-tuning, and disease category-based balanced instruction-tuning, respectively. Note: P2Med-MLLM: Medical Multimodal Large Language Model for Pediatric Pneumonia. CT: Computed Tomography.}
  \label{ablation_study_stage_modality}
\end{figure}

To investigate the impact of different stages and modalities, we provided a thorough ablation study of the Medical Multimodal Large Language Model for Pediatric Pneumonia~(P2Med-MLLM) by removing single stage or modality. The results were shown in Table \ref{ablation_study_stage_modality_main} and Fig.~\ref{ablation_study_stage_modality}. 

First, we investigated the impact of different training stages on the most critical evaluation components of each task, specifically impression or diagnosis results~(columns Full and A-C in Table \ref{ablation_study_stage_modality_main}). We found that each stage contributed to performance improvement, demonstrating the significance of medical knowledge infusion pre-training, task type-based balanced instruction tuning, and disease category-based balanced instruction tuning in multi-task clinical decision supports. Specifically, the importance of the three stages, in descending order, were: stage 1, stage 2, and stage 3, respectively.

In addition to the three-stage training strategy, we also evaluated the impact of different modalities, that is, plain text and image-text data. By comparing column Full with columns D and E in Table \ref{ablation_study_stage_modality_main}, respectively, we observed that removing either modality adversely affected the performance of the other~(at least 0.6 on average). These results suggested that tasks involving both modalities were mutually beneficial to some extent. Notably, image-text tasks had a more significant influence on plain text tasks.

Next, as shown in Fig.~\ref{ablation_study_stage_modality}, we explored the performance of P2Med-MLLM across different stages and modalities for all evaluation components of each task. Compared to incomplete stages and modalities, P2Med-MLLM demonstrated significant advantages. Specifically, P2Med-MLLM outperformed others in 4 out of the 6 most crucial evaluation components and in 9 out of 16 evaluation components overall. These observations suggested that while P2Med-MLLM achieved the best results, some evaluation components, such as diagnostic basis and treatment plan, still showed room for improvement. We believed the reason behind this was that the ground truth for these open-ended evaluation components was inherently diverse, making automatic scoring with language models more challenging. Therefore, we focused primarily on the most crucial and standardized evaluation components of each task.

\subsection{Impact of Different Tasks in P2Med-MLLM}
\begin{figure}[htbp]
\centering
\begin{subfigure}[b]{\columnwidth}
    \begin{tikzpicture}
      \node[anchor=south west,inner sep=0] (image) at (0,0) {\includegraphics[width=\columnwidth]{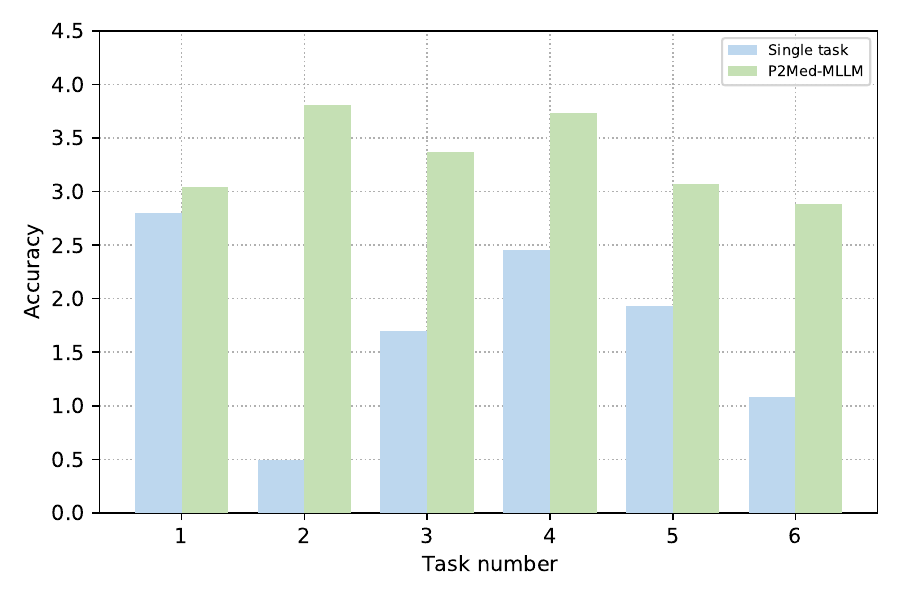}};
      \node[anchor=north west] at (image.north west) {\textbf{a}}; 
    \end{tikzpicture}
    \caption{}
    \label{ablation_study_task_accuracy}
\end{subfigure}
\vspace{-3.0em}

\begin{subfigure}[b]{\columnwidth}
    \begin{tikzpicture}
      \node[anchor=south west,inner sep=0] (image) at (0,0) {\includegraphics[width=\columnwidth]{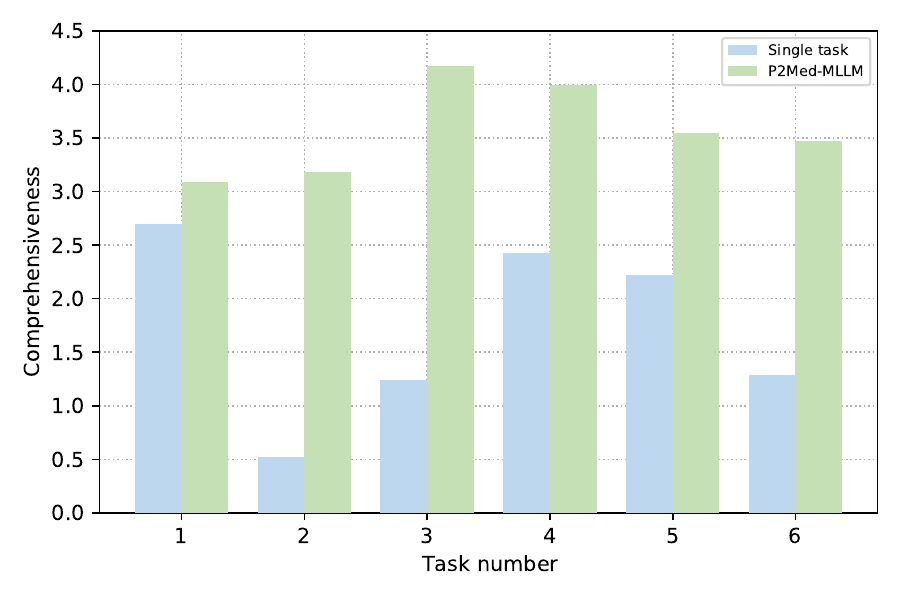}};
      \node[anchor=north west] at (image.north west) {\textbf{b}}; 
    \end{tikzpicture}
    \caption{}
    \label{ablation_study_task_comprehensiveness}
\end{subfigure}
\vspace{-3.0em}

\caption{\textbf{Performance comparison between multiple single-task dedicated networks and a unified network trained jointly on multiple tasks~(P2Med-MLLM).} Accuracy~(\textbf{a}) and Comprehensiveness~(\textbf{b}) scores of impression or diagnosis results were reported, representing the key metrics of evaluation. Task 1 to task 6 represented radiology report generation~(X-ray), radiology report generation~(CT), outpatient medical record generation, first disease course record generation, attending physician's first ward round record generation, and chief physician's first ward round record generation, respectively. Note: P2Med-MLLM: Medical Multimodal Large Language Model for Pediatric Pneumonia. CT: Computed Tomography.}
\label{ablation_study_task}
\vspace{-1.5em}
\end{figure}

Traditional methods typically involved training a network on a subset for a specific medical task. While intuitive, such a training strategy significantly increased computational complexity. To demonstrate the effectiveness of P2Med-MLLM trained jointly on multiple tasks, we compared it with multiple single-task dedicated networks on the most crucial evaluation components using accuracy and comprehensiveness metrics. As shown in Fig.~\ref{ablation_study_task}, joint training by P2Med-MLLM yielded substantial performance improvements, particularly in tasks such as radiology report generation~(Computed Tomography, CT), outpatient medical record generation, and chief physician’s first ward round record generation. We utilized a generative network to unify all tasks, and this flexible structure ensured performance while easily extending to new tasks. It was valuable in the real world for assisting clinicians in completing multiple tasks.

\subsection{Impact of Different Conversation Forms in P2Med-MLLM}
We found that during a single outpatient or inpatient visit for each patient, there may be multiple scans for the same imaging modality, reflecting changes in the patient's condition. Thus, for the radiology report generation task, we constructed a multi-round conversation using all scans of the same imaging modality from a single visit, arranged in chronological order. Using X-ray scans as an example, we compared the performance of models with and without multi-round conversations. As shown in Table \ref{ablation_study_image}, ``w/o MRC" indicated treating each scan as an independent single-round conversation. Although P2Med-MLLM and ``w/o MRC" achieved comparable results in the evaluation component of findings, adopting multi-round conversations showed a notable advantage in the most crucial evaluation component of impression, exceeding in both accuracy and comprehensiveness metrics by at least 0.37. This demonstrated that the temporal information was critical to perform radiological diagnosis.

\begin{table}[h]
\begin{threeparttable}
\caption{\textbf{An ablation study of P2Med-MLLM in radiology report generation~(X-ray) task.} The metrics presented reflected the average scores across all test samples in radiology report generation~(X-ray) task, with 95\% confidence intervals in parentheses. The best results were bolded.}
\label{ablation_study_image}
\renewcommand\arraystretch{1.5}
\setlength{\tabcolsep}{10.4pt}
\begin{tabular}{@{}llll@{}}
\hline
\multirow{2.4}{*}{Method} & Findings                                                             & \multicolumn{2}{l}{Impression}                                                                                                              \\ \cmidrule(r){2-2} \cmidrule(r){3-4}
                        & Accuracy                                                             & Accuracy                                                             & Comprehensiveness                                                    \\ \hline
P2Med-MLLM              & \textbf{\begin{tabular}[c]{@{}l@{}}3.54\\ (3.41, 3.67)\end{tabular}} & \textbf{\begin{tabular}[c]{@{}l@{}}3.04\\ (2.75, 3.33)\end{tabular}} & \textbf{\begin{tabular}[c]{@{}l@{}}3.09\\ (2.83, 3.35)\end{tabular}} \\ \hline
W/o MRC                 & \begin{tabular}[c]{@{}l@{}}3.49\\ (3.37, 3.61)\end{tabular}          & \begin{tabular}[c]{@{}l@{}}2.67\\ (2.42, 2.92)\end{tabular}          & \begin{tabular}[c]{@{}l@{}}2.53\\ (2.35, 2.71)\end{tabular}          \\ \hline
\end{tabular}
\begin{tablenotes}[para,flushleft]
\item Note: P2Med-MLLM: Medical Multimodal Large Language Model for Pediatric Pneumonia. MRC: Multi-Round Conversations.
\end{tablenotes}
\end{threeparttable}
\vspace{-1.5em}
\end{table}

\subsection{Impact of Different Large Language Models in P2Med-MLLM}
In this subsection, we explored different Large Language Models~(LLMs) in P2Med-MLLM using the Medical Multimodal Benchmark for Pediatric Pneumonia~(P2Med-MBench). Specifically, we compared models based on Baichuan 2~\cite{yang2023baichuan} and Chinese-LLAMA-2~\cite{cui2023efficient}. The results in Table \ref{ablation_llm} showed that the Chinese-LLAMA-2-based model significantly outperformed Baichuan 2-based model on average and most tasks, except for the outpatient medical record generation task. Therefore, we chose Chinese-LLAMA-2 as the Large Language Model~(LLM) for P2Med-MLLM.

\begin{table*}[h]
\begin{threeparttable}
\caption{\textbf{An ablation study of P2Med-MLLM with different LLMs.} Accuracy and Comprehensiveness scores of impression or diagnosis results were reported, representing the key metrics of evaluation. The metrics presented reflected the average scores across all test samples, with 95\% confidence intervals in parentheses. The best results were bolded. Task 1 to task 6 represented radiology report generation~(X-ray), radiology report generation~(CT), outpatient medical record generation, first disease course record generation, attending physician's first ward round record generation, and chief physician's first ward round record generation, respectively.}
\label{ablation_llm}
\renewcommand\arraystretch{1.5}
\setlength{\tabcolsep}{4.45pt}
\tiny
\begin{tabular}{lllllllllllllll}
\hline
\multirow{2}{*}{Method} & \multicolumn{2}{l}{Task 1}                                                                                                                  & \multicolumn{2}{l}{Task 2}                                                                                                                  & \multicolumn{2}{l}{Task 3}                                                                                                                  & \multicolumn{2}{l}{Task 4}                                                                                                                  & \multicolumn{2}{l}{Task 5}                                                                                                                  & \multicolumn{2}{l}{Task 6}                                                                                                                  & \multicolumn{2}{l}{Average}                                                                                                                     \\ \cmidrule(r){2-3} \cmidrule(r){4-5} \cmidrule(r){6-7} \cmidrule(r){8-9} \cmidrule(r){10-11} \cmidrule(r){12-13} \cmidrule(r){14-15} 
                        & Acc                                                                  & Comp                                                                 & Acc                                                                  & Comp                                                                 & Acc                                                                  & Comp                                                                 & Acc                                                                  & Comp                                                                 & Acc                                                                  & Comp                                                                 & Acc                                                                  & Comp                                                                 & Acc                                                                  & Comp                                                                 \\ \hline
Baichuan 2~\cite{yang2023baichuan}              & \begin{tabular}[c]{@{}l@{}}2.87\\ (2.62, 3.12)\end{tabular}          & \begin{tabular}[c]{@{}l@{}}2.78\\ (2.53, 3.03)\end{tabular}          & \begin{tabular}[c]{@{}l@{}}2.89\\ (2.58, 3.20)\end{tabular}          & \begin{tabular}[c]{@{}l@{}}2.56\\ (2.41, 2.71)\end{tabular}          & \textbf{\begin{tabular}[c]{@{}l@{}}3.41\\ (3.19, 3.63)\end{tabular}} & \textbf{\begin{tabular}[c]{@{}l@{}}4.19\\ (3.99, 4.39)\end{tabular}} & \begin{tabular}[c]{@{}l@{}}2.55\\ (2.12, 2.98)\end{tabular}          & \begin{tabular}[c]{@{}l@{}}2.62\\ (2.18, 3.06)\end{tabular}          & \begin{tabular}[c]{@{}l@{}}2.78\\ (2.42, 3.14)\end{tabular}          & \begin{tabular}[c]{@{}l@{}}3.15\\ (2.75, 3.55)\end{tabular}          & \begin{tabular}[c]{@{}l@{}}2.38\\ (2.02, 2.74)\end{tabular}          & \begin{tabular}[c]{@{}l@{}}2.77\\ (2.37, 3.17)\end{tabular}          & \begin{tabular}[c]{@{}l@{}}2.81\\ (2.67, 2.95)\end{tabular}          & \begin{tabular}[c]{@{}l@{}}3.01\\ (2.87, 3.15)\end{tabular}          \\ \hline
Chinese-LLaMA-2~\cite{cui2023efficient}         & \textbf{\begin{tabular}[c]{@{}l@{}}3.04\\ (2.75, 3.33)\end{tabular}} & \textbf{\begin{tabular}[c]{@{}l@{}}3.09\\ (2.83, 3.35)\end{tabular}} & \textbf{\begin{tabular}[c]{@{}l@{}}3.81\\ (3.50, 4.12)\end{tabular}} & \textbf{\begin{tabular}[c]{@{}l@{}}3.18\\ (2.97, 3.39)\end{tabular}} & \begin{tabular}[c]{@{}l@{}}3.37\\ (3.20, 3.54)\end{tabular}          & \begin{tabular}[c]{@{}l@{}}4.17\\ (4.04, 4.30)\end{tabular}          & \textbf{\begin{tabular}[c]{@{}l@{}}3.73\\ (3.47, 3.99)\end{tabular}} & \textbf{\begin{tabular}[c]{@{}l@{}}3.99\\ (3.72, 4.26)\end{tabular}} & \textbf{\begin{tabular}[c]{@{}l@{}}3.07\\ (2.80, 3.34)\end{tabular}} & \textbf{\begin{tabular}[c]{@{}l@{}}3.55\\ (3.24, 3.86)\end{tabular}} & \textbf{\begin{tabular}[c]{@{}l@{}}2.88\\ (2.61, 3.15)\end{tabular}} & \textbf{\begin{tabular}[c]{@{}l@{}}3.47\\ (3.15, 3.79)\end{tabular}} & \textbf{\begin{tabular}[c]{@{}l@{}}3.32\\ (3.21, 3.43)\end{tabular}} & \textbf{\begin{tabular}[c]{@{}l@{}}3.57\\ (3.46, 3.68)\end{tabular}} \\ \hline
\end{tabular}
\begin{tablenotes}[para,flushleft]
\item Note: P2Med-MLLM: Medical Multimodal Large Language Model for Pediatric Pneumonia. LLMs: Large Language Models. CT: Computed Tomography. Acc: Accuracy. Comp: Comprehensiveness.
\end{tablenotes}
\end{threeparttable}
\end{table*}
\section{Outlook}

Multimodal Large Language Models~(MLLMs) have brought substantial advancements in the healthcare field. In this study, we preliminarily explored and demonstrated the feasibility of securely and effectively training and deploying a MLLM on private hospital data, specifically focusing on real clinical scenarios involving patients with a primary diagnosis of pediatric pneumonia. Our work encompassed the entire process, from data collection and cleaning to model construction and evaluation, offering a valuable reference for researchers in the interdiscipline of artificial intelligence for medicine. We built the largest Chinese Medical Multimodal Dataset for Pediatric Pneumonia~(P2Med-MD) so far. Different from previous efforts, the Medical Multimodal Large Language Model for Pediatric Pneumonia~(P2Med-MLLM) employed a unified framework that supported both pure text data~(outpatient and inpatient records) and temporally sequenced, interleaved 2D or 3D medical images alongside radiology reports, aligning more closely with clinical practice. P2Med-MLLM could potentially serve as a clinical assistant, helping doctors enhance diagnostic and treatment efficiency, providing personalized recommendations for pediatric pneumonia patients, and optimizing clinical workflows.

Despite the progress achieved in our research, there are several limitations. Firstly, for complicated and open-ended clinical tasks, such as generating diagnostic bases and treatment plans in medical records, the performance of P2Med-MLLM still falls short of real clinical applications. Additionally, the automatic scoring system lacks robustness, highlighting the need for more objective evaluation metrics. Secondly, this study only includes patients primarily diagnosed with pediatric pneumonia. Future work could extend the objects to cover all respiratory diseases, or even the entire spectrum of general medicine across all age groups. Thirdly, the study is limited to a single-center cohort, and data collection from multiple healthcare institutions and countries would enhance diversity and generalizability. Lastly, as this is a retrospective study, future research could explore prospective studies.
\section{Methods}

In this section, we provided a detailed description of our self-built dataset, the medical multimodal Large Language Model~(LLM), and the implementation details.

\subsection{Medical Multimodal Dataset for Pediatric Pneumonia~(P2Med-MD)}
Currently, the medical domain faces a significant shortfall in multimodal datasets that accurately reflect real-world clinical scenarios, a crucial element for training a practical medical multimodal LLM. To bridge this gap, we constructed a high-quality, large-scale Chinese \textbf{Med}ical \textbf{M}ultimodal \textbf{D}ataset for \textbf{P}ediatric \textbf{P}neumonia~(P2Med-MD), through human-machine interaction. P2Med-MD focused on pediatric patients with a primary diagnosis of pneumonia. Here, we started by providing an overview of P2Med-MD in Sec~\ref{overview}. It consisted of three sets, medical knowledge infusion, task type-based balanced sampling, and disease category-based balanced sampling, corresponding to Sec~\ref{stage1_data}, ~\ref{stage2_data}, and ~\ref{stage3_data}, respectively. These parts were utilized for different training stages described in Sec ~\ref{model}. In Sec~\ref{benchmark}, we introduced a new \textbf{Med}ical \textbf{M}ultimodal \textbf{Bench}mark for \textbf{P}ediatric \textbf{P}neumonia, termed P2Med-MBench, which encompasses six tasks, \textit{e.g.}, radiology report generation~(X-ray), radiology report generation~(Computed Tomography, CT), outpatient medical record generation, first disease course record generation, attending physician's first ward round record generation, and chief physician's first ward round record generation. These tasks were designed to monitor the development of the medical multimodal LLM.

\subsubsection{Overview\label{overview}}
\begin{figure*}[htbp]
\centering
\begin{subfigure}[t]{0.31\textwidth}
    \begin{tikzpicture}
      \node[anchor=south west,inner sep=0] (image) at (0,0) {\includegraphics[height=5cm, keepaspectratio]{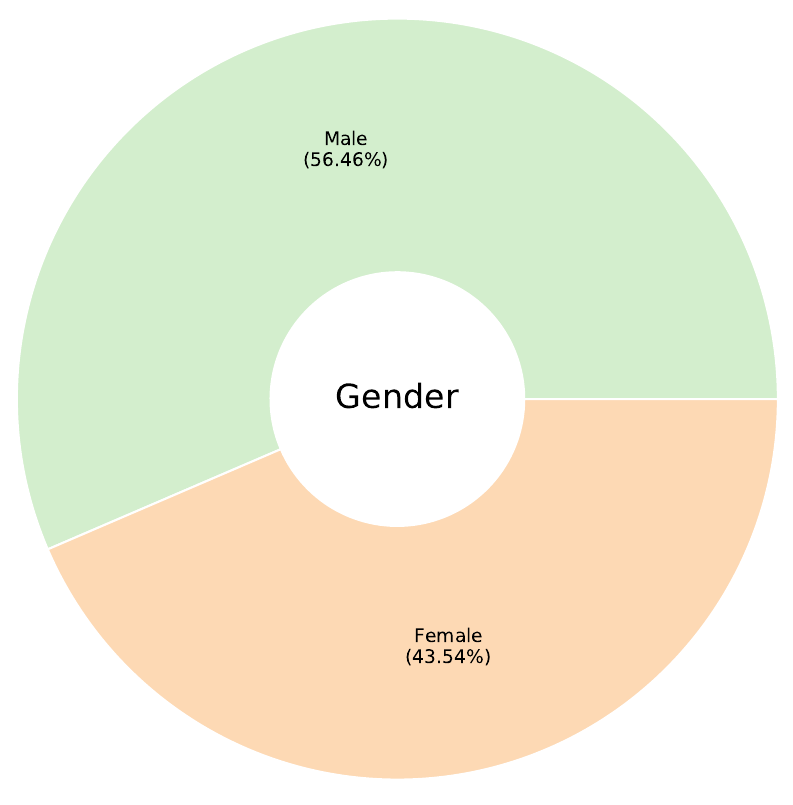}};
      \node[anchor=north west] at (image.north west) {\textbf{a}}; 
    \end{tikzpicture}
    \caption{}
    \label{dataset_distribution_a}
  \end{subfigure}
  \begin{subfigure}[t]{0.31\textwidth}
    \begin{tikzpicture}
      \node[anchor=south west,inner sep=0] (image) at (0,0) {\includegraphics[height=5cm, keepaspectratio]{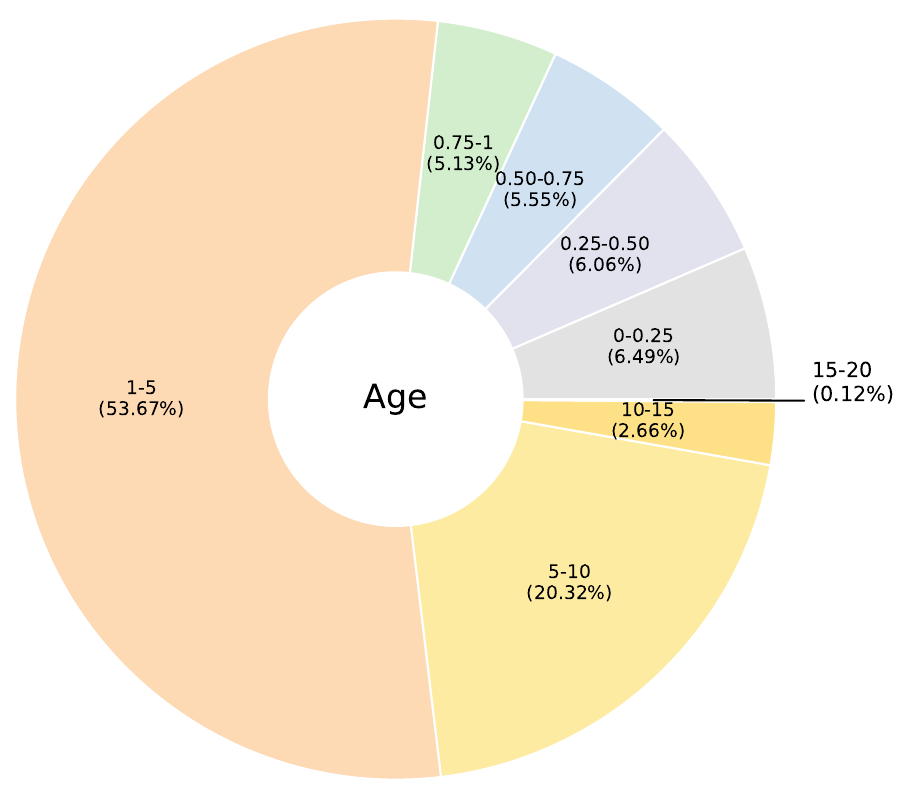}};
      \node[anchor=north west] at (image.north west) {\textbf{b}}; 
    \end{tikzpicture}
    \caption{}
    \label{dataset_distribution_b}
  \end{subfigure}
  \begin{subfigure}[t]{0.31\textwidth}
    \begin{tikzpicture}
      \node[anchor=south west,inner sep=0] (image) at (0,0) {\includegraphics[height=5cm, keepaspectratio]{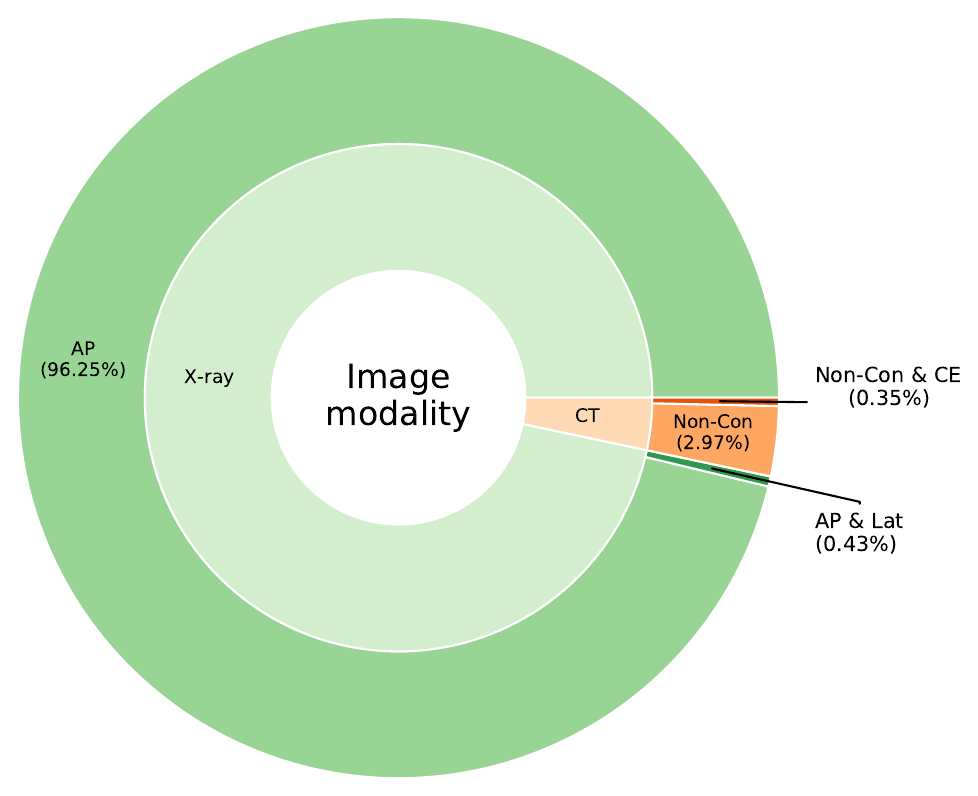}};
      \node[anchor=north west] at (image.north west) {\textbf{c}}; 
    \end{tikzpicture}
    \caption{}
    \label{dataset_distribution_c}
  \end{subfigure}
  \vspace{-1.0em}
  \caption{\textbf{The data statistics of P2Med-MD.} (\textbf{a}) The distribution of gender. (\textbf{b}) The distribution of age~(in years). (\textbf{c}) The diversity in image modalities. The collected dataset encompassed approximately 70K radiology images, spanning two modalities of varying dimensions: 2D for X-ray and 3D for CT scans. Note: P2Med-MD: Medical Multimodal Dataset for Pediatric Pneumonia. CT: Computed Tomography. AP: AnteroPosterior view. Lat: Lateral view. Non-Con: Non-Contrast series. CE: Contrast-Enhanced series.}
  \label{dataset_distribution}
\end{figure*}
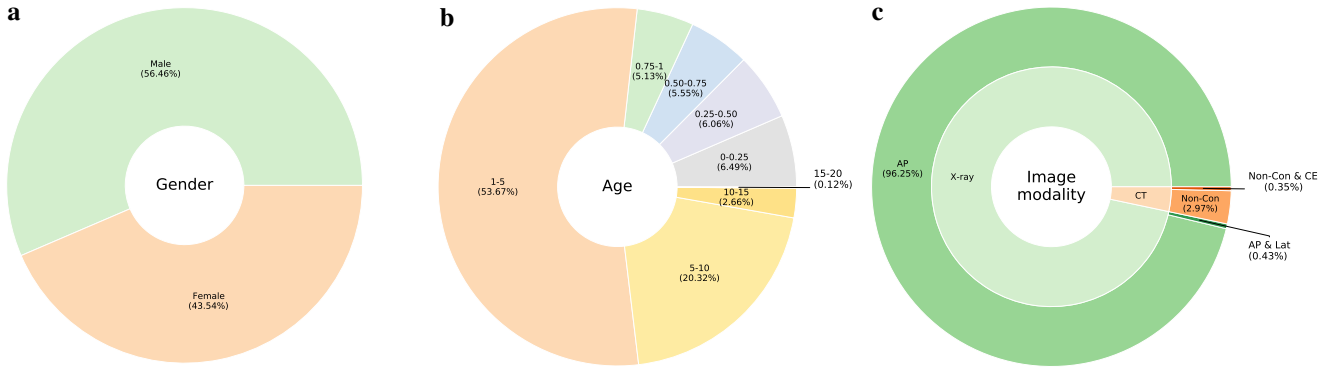

The study was approved by the Ethics Committee of Children’s Hospital, Fudan University~(2022-307A, approved November 22, 2022). For participants admitted before November 22, 2022, informed consent was waived; for those admitted on or after November 22, 2022, informed consent was obtained. In this retrospective study, we collected the outpatient information of 163,999 patients and the inpatient information of 8,684 patients who were admitted to Children’s Hospital of Fudan University between August 26, 2016 to November 1, 2023. Outpatient information included outpatient medical records, and chest X-ray and CT scans along with corresponding radiology reports. Inpatient information comprised three-level round records formed by first disease course records, attending physician's first ward round records, and chief physician's first ward round records, as well as chest X-ray and CT scans with corresponding radiology reports. The built dataset altogether contained 67,616 chest X-ray examinations and 2,321 chest CT examinations along with their respective radiology reports, 684,758 outpatient medical records, 9,180 first disease course records, 9,993 attending physician's first ward round records, and 6,426 chief physician's first ward round records. More details were given in Table \ref{P2Med-MD}. Fig.~\ref{dataset_distribution} illustrated the distribution of gender, age, and image modalities in P2Med-MD.

\begin{table*}[htbp]
\centering
\caption{\textbf{Description of the P2Med-MD.} Stage 1 to stage 3 represented medical knowledge infusion pre-training, task type-based balanced instruction-tuning, and disease category-based balanced instruction-tuning, respectively.}
\label{P2Med-MD}
\renewcommand\arraystretch{2.0}
\setlength{\tabcolsep}{11.4pt}
\begin{tabular}{lllllll}
\hline
Task number & Task description                                           & Total   & Stage 1 & Stage 2 & Stage 3 & Test set \\ \hline
1             & Radiology report generation~(X-ray)                   & 67,616  & 67,495  & 67,495  & 14,461  & 121      \\ \hline
2             & Radiology report generation~(CT)                      & 2,321   & 2,200   & 2,200   & 2,200   & 121      \\ \hline
3             & Outpatient medical record generation                       & 684,758 & 407,185 & 23,560  & 10,200  & 100      \\ \hline
4             & First disease course record generation                     & 9,180   & -       & 9,080   & 3,255   & 100      \\ \hline
5             & Attending physician's first ward round   record generation & 9,993   & -       & 9,893   & 4,713   & 100      \\ \hline
6             & Chief physician's first ward round record generation       & 6,426   & -       & 6,326   & 3,828   & 100      \\ \hline
\end{tabular}
\begin{tablenotes}[para,flushleft]
\item Note: P2Med-MD: Medical Multimodal Dataset for Pediatric Pneumonia. CT: Computed Tomography.
\end{tablenotes}
\end{table*}

\begin{figure*}[htbp]
  \centering
  \begin{subfigure}[t]{0.32\textwidth}
    \begin{tikzpicture}
      \node[anchor=south west,inner sep=0] (image) at (0,0) {\includegraphics[height=4.5cm, width=\linewidth]{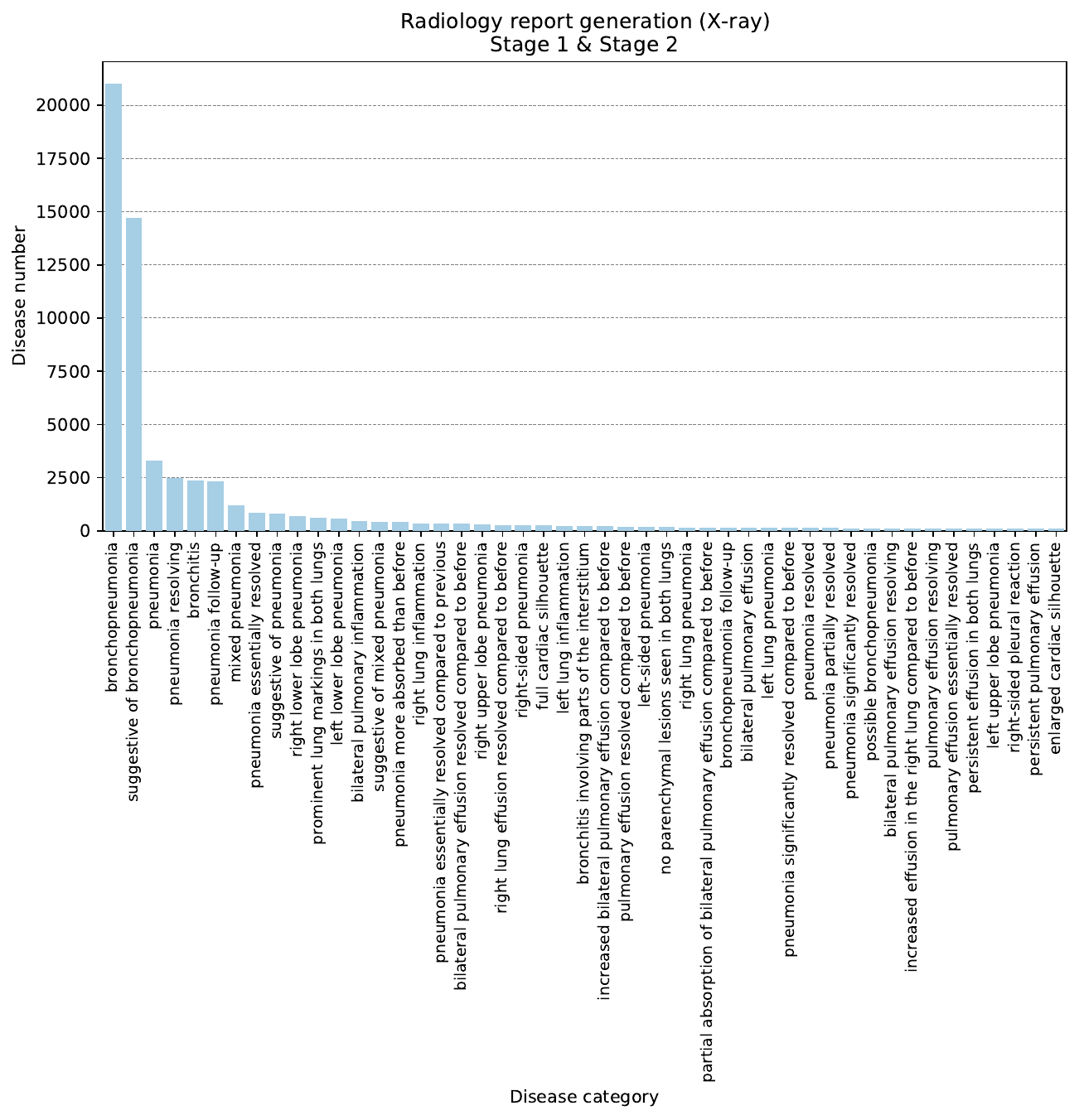}};
      \node[anchor=north west, yshift=1pt] at (image.north west) {\textbf{a}}; 
    \end{tikzpicture}
    \caption{}
    \label{disease_distribution_a}
  \end{subfigure}%
  \hfill
  \begin{subfigure}[t]{0.32\textwidth}
    \begin{tikzpicture}
      \node[anchor=south west,inner sep=0] (image) at (0,0) {\includegraphics[height=4.5cm, width=\linewidth]{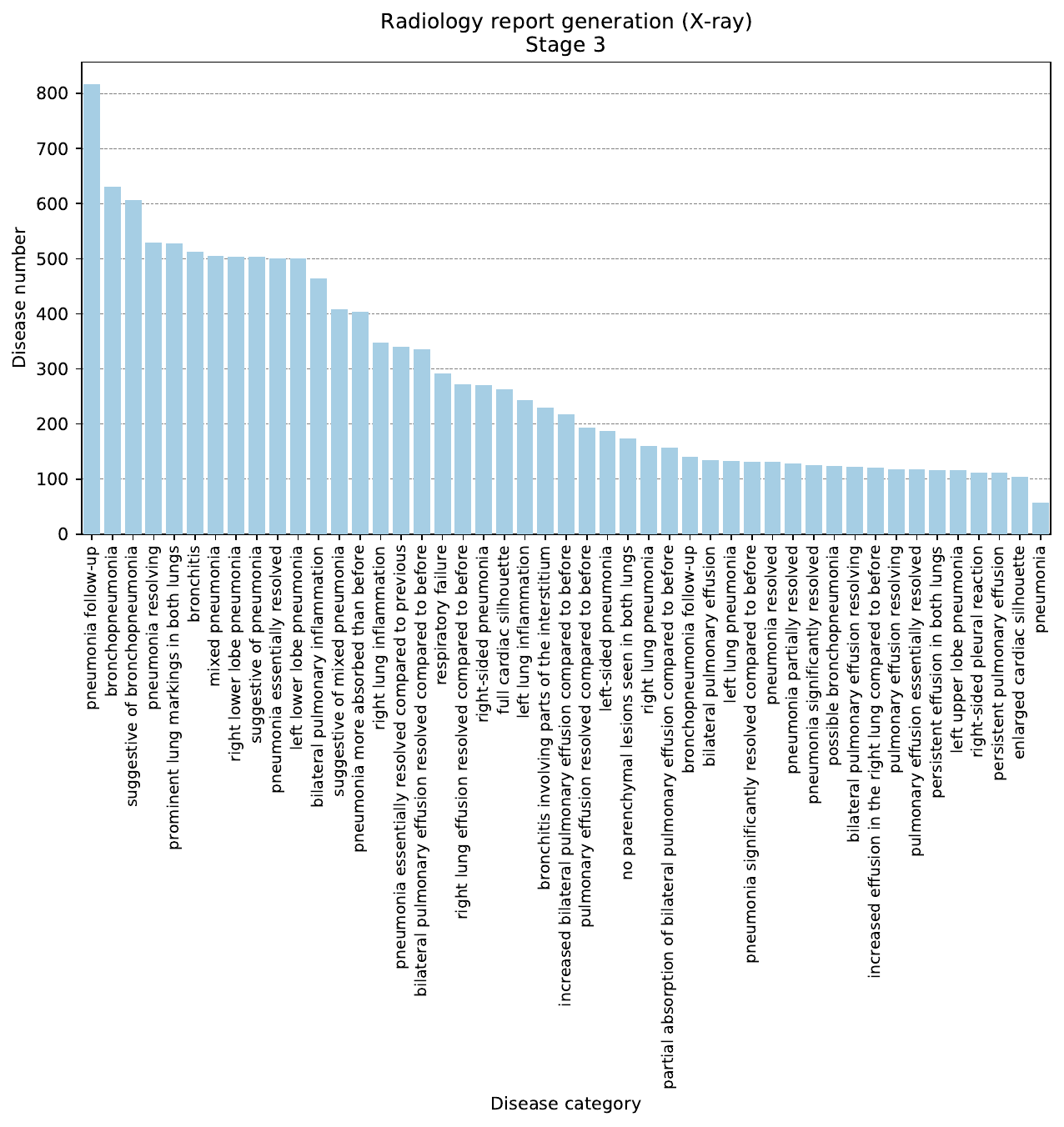}};
      \node[anchor=north west, yshift=1pt] at (image.north west) {\textbf{b}}; 
    \end{tikzpicture}
    \caption{}
    \label{disease_distribution_b}
  \end{subfigure}%
  \hfill
  \begin{subfigure}[t]{0.32\textwidth}
    \begin{tikzpicture}
      \node[anchor=south west,inner sep=0] (image) at (0,0) {\includegraphics[height=4.5cm, width=\linewidth]{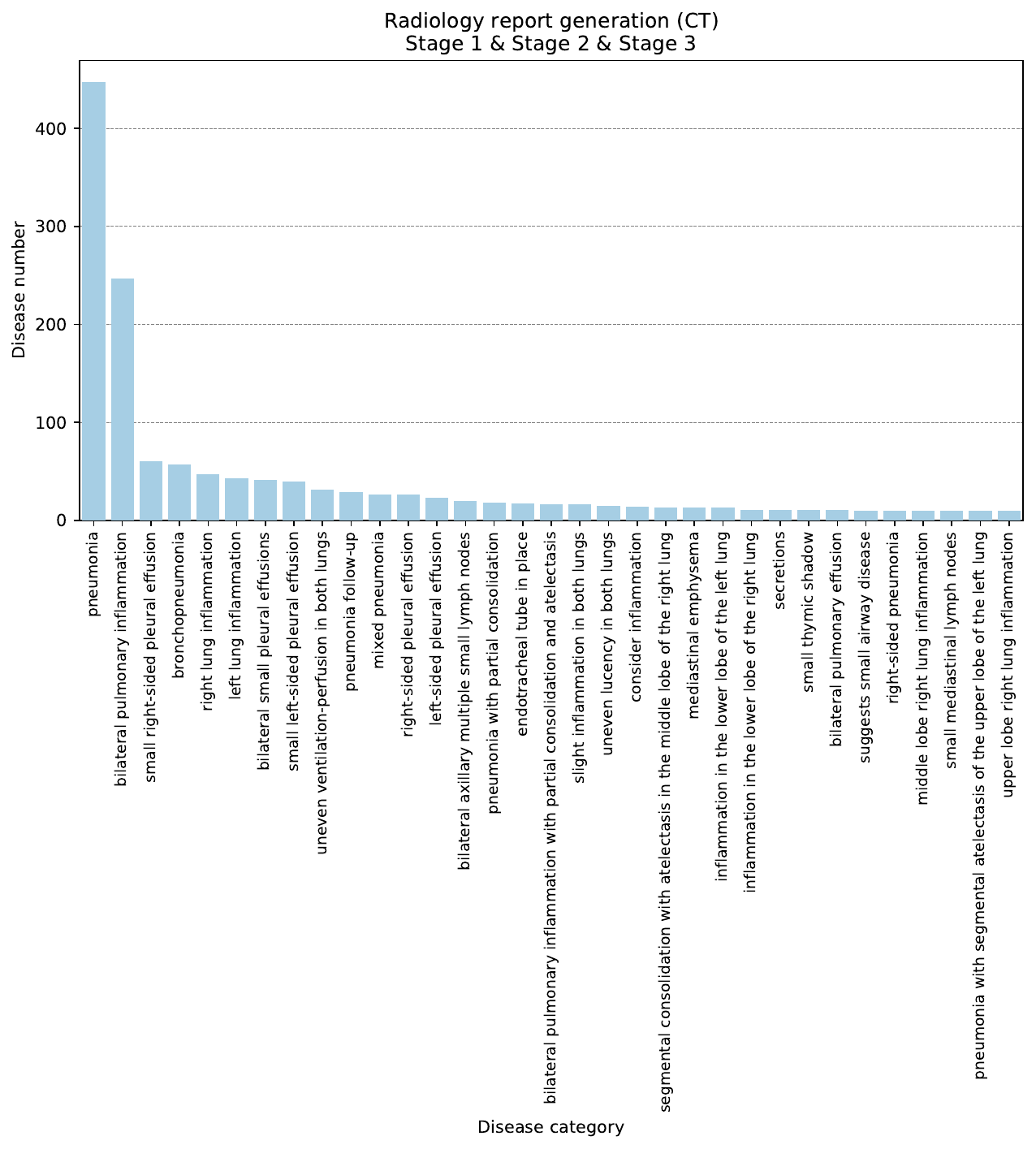}};
      \node[anchor=north west, yshift=1pt] at (image.north west) {\textbf{c}}; 
    \end{tikzpicture}
    \caption{}
    \label{disease_distribution_c}
  \end{subfigure}
  \vspace{-0.8em}
  
  \begin{subfigure}[b]{0.48\textwidth}
    \begin{tikzpicture}
      \node[anchor=south west,inner sep=0] (image) at (0,0) {\includegraphics[height=4.5cm, width=\linewidth]{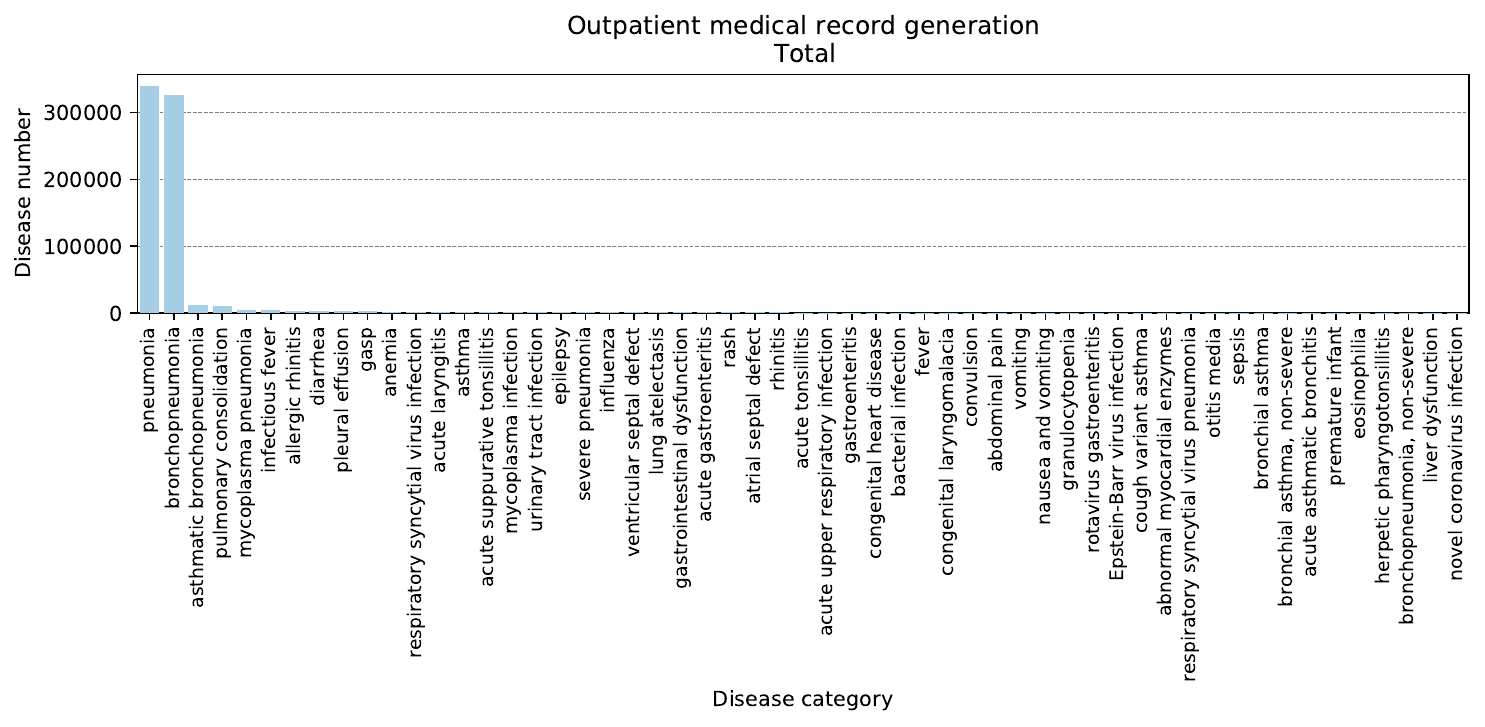}};
      \node[anchor=north west] at (image.north west) {\textbf{d}}; 
    \end{tikzpicture}
    \caption{}
    \label{disease_distribution_d}
  \end{subfigure}%
  \hfill
  \begin{subfigure}[b]{0.48\textwidth}
    \begin{tikzpicture}
      \node[anchor=south west,inner sep=0] (image) at (0,0) {\includegraphics[height=4.5cm, width=\linewidth]{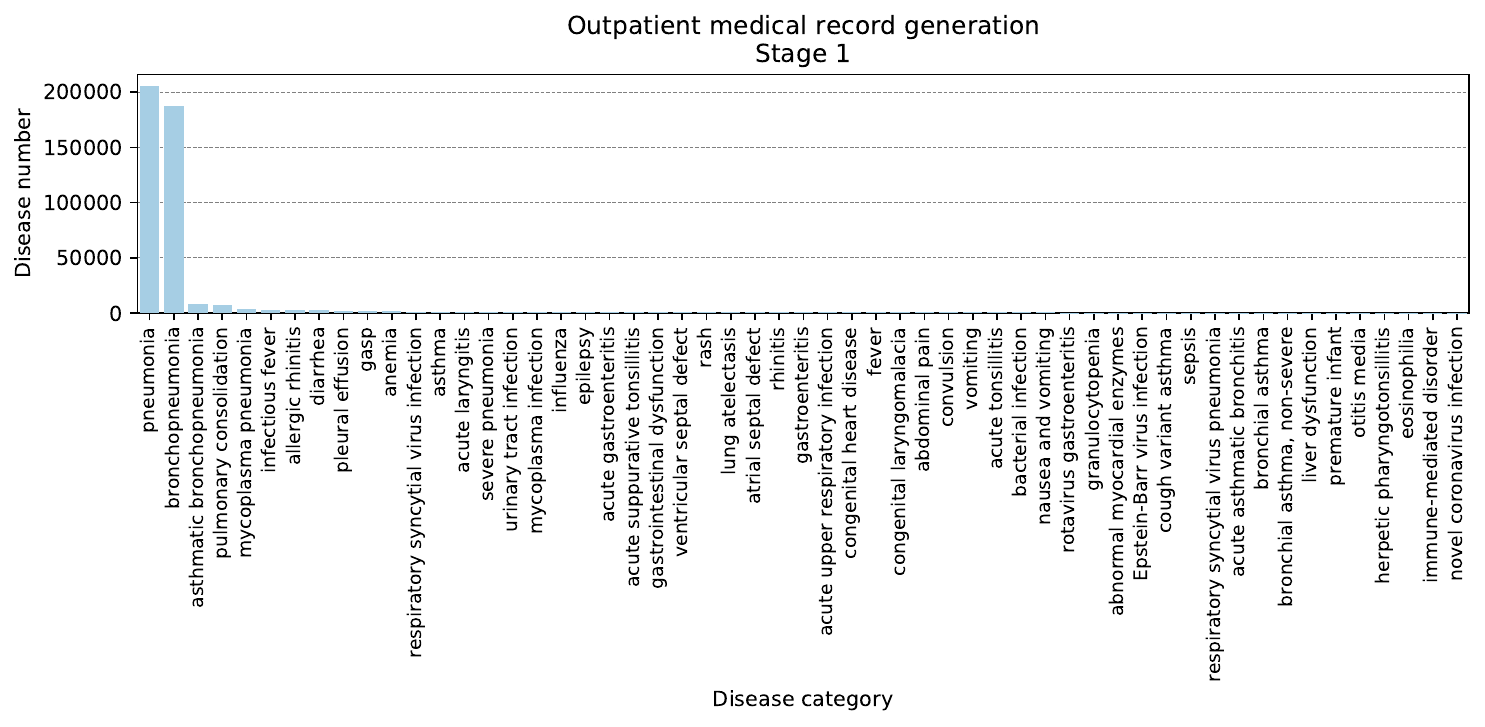}};
      \node[anchor=north west] at (image.north west) {\textbf{e}}; 
    \end{tikzpicture}
    \caption{}
    \label{disease_distribution_e}
  \end{subfigure}
  \vspace{-0.8em}

  \begin{subfigure}[b]{0.48\textwidth}
    \begin{tikzpicture}
      \node[anchor=south west,inner sep=0] (image) at (0,0) {\includegraphics[height=4.5cm, width=\linewidth]{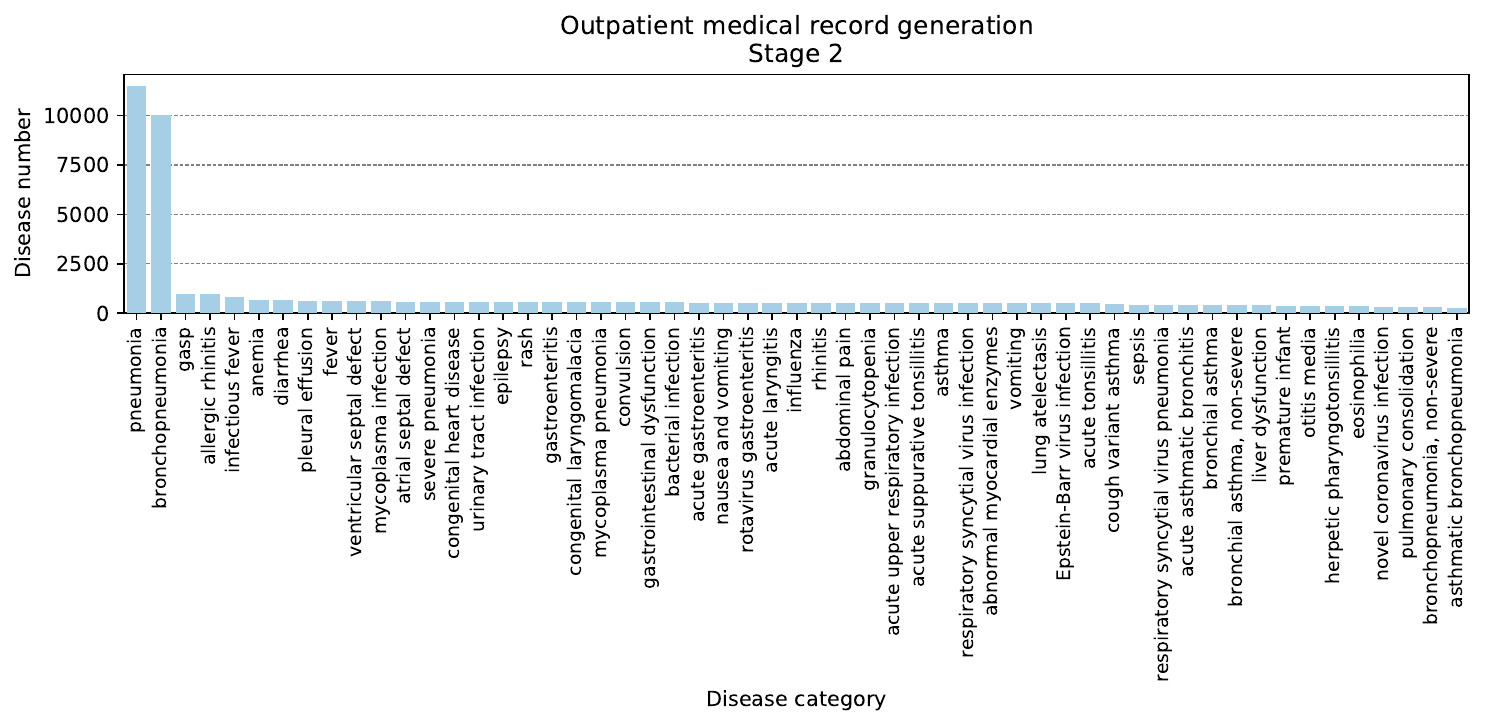}};
      \node[anchor=north west] at (image.north west) {\textbf{f}}; 
    \end{tikzpicture}
    \caption{}
    \label{disease_distribution_f}
  \end{subfigure}%
  \hfill
  \begin{subfigure}[b]{0.48\textwidth}
    \begin{tikzpicture}
      \node[anchor=south west,inner sep=0] (image) at (0,0) {\includegraphics[height=4.5cm, width=\linewidth]{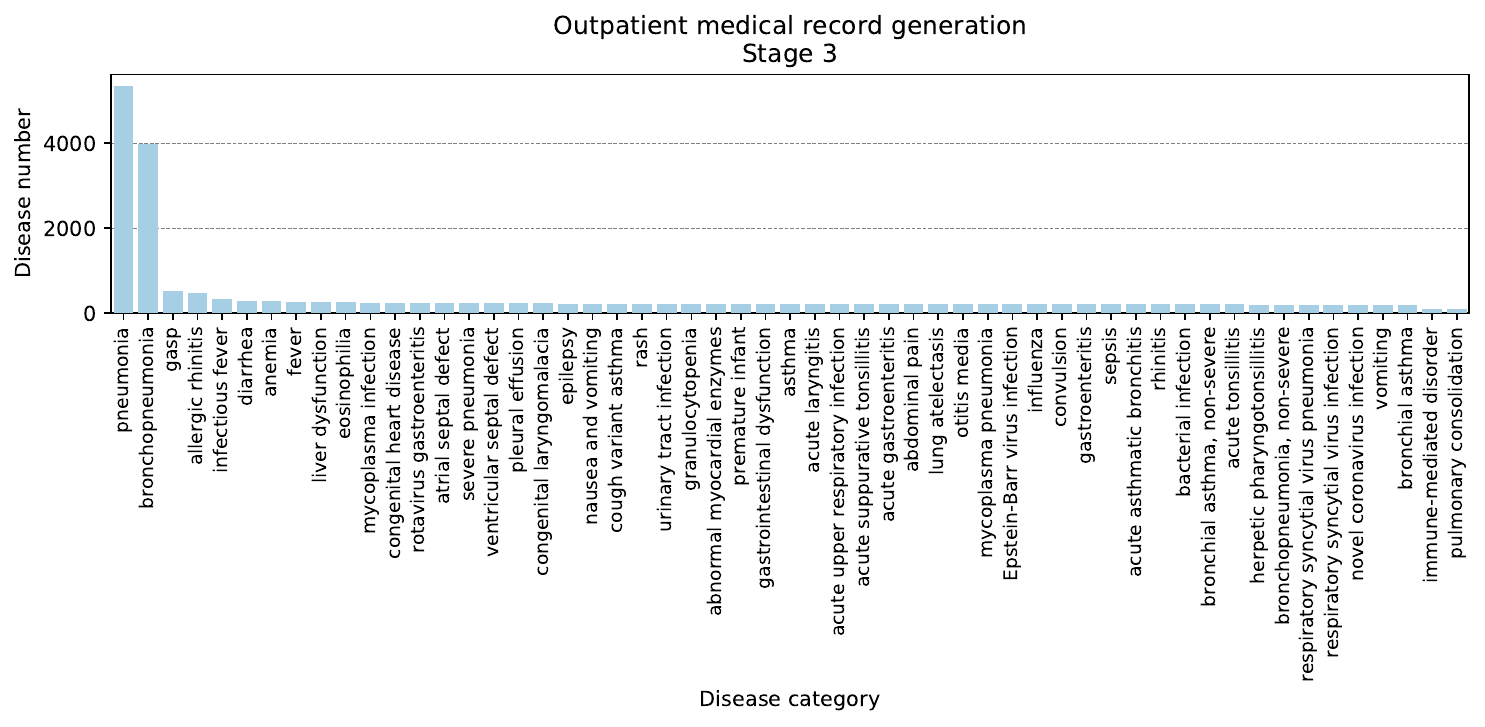}};
      \node[anchor=north west] at (image.north west) {\textbf{g}}; 
    \end{tikzpicture}
    \caption{}
    \label{disease_distribution_g}
  \end{subfigure}
  \vspace{-0.8em}
  
  \begin{subfigure}[b]{0.32\textwidth}
    \begin{tikzpicture}
      \node[anchor=south west,inner sep=0] (image) at (0,0) {\includegraphics[height=4.5cm, width=\linewidth]{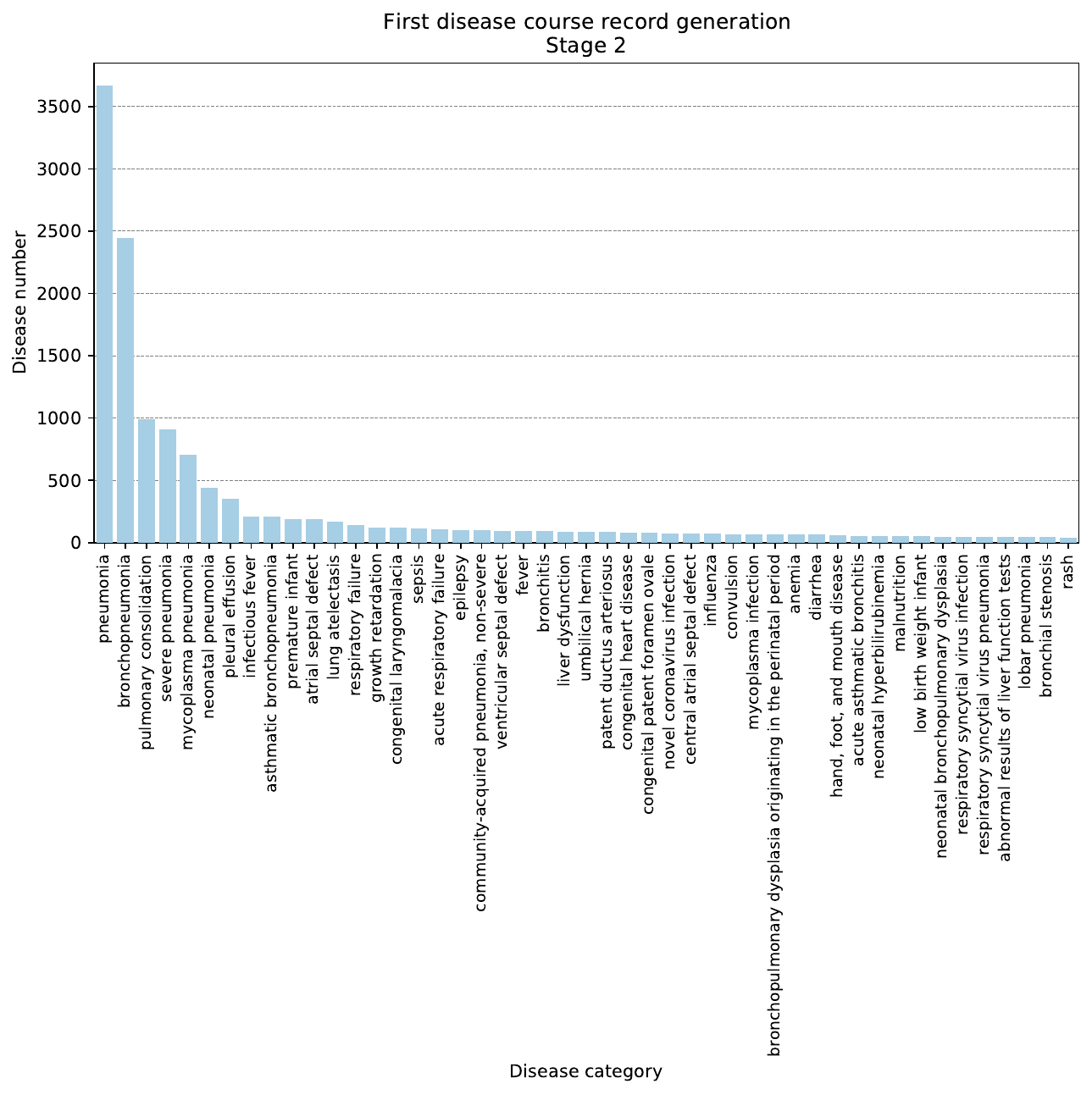}};
      \node[anchor=north west, yshift=4pt] at (image.north west) {\textbf{h}}; 
    \end{tikzpicture}
    \caption{}
    \label{disease_distribution_h}
  \end{subfigure}%
  \hfill
  \begin{subfigure}[b]{0.32\textwidth}
    \begin{tikzpicture}
      \node[anchor=south west,inner sep=0] (image) at (0,0) {\includegraphics[height=4.5cm, width=\linewidth]{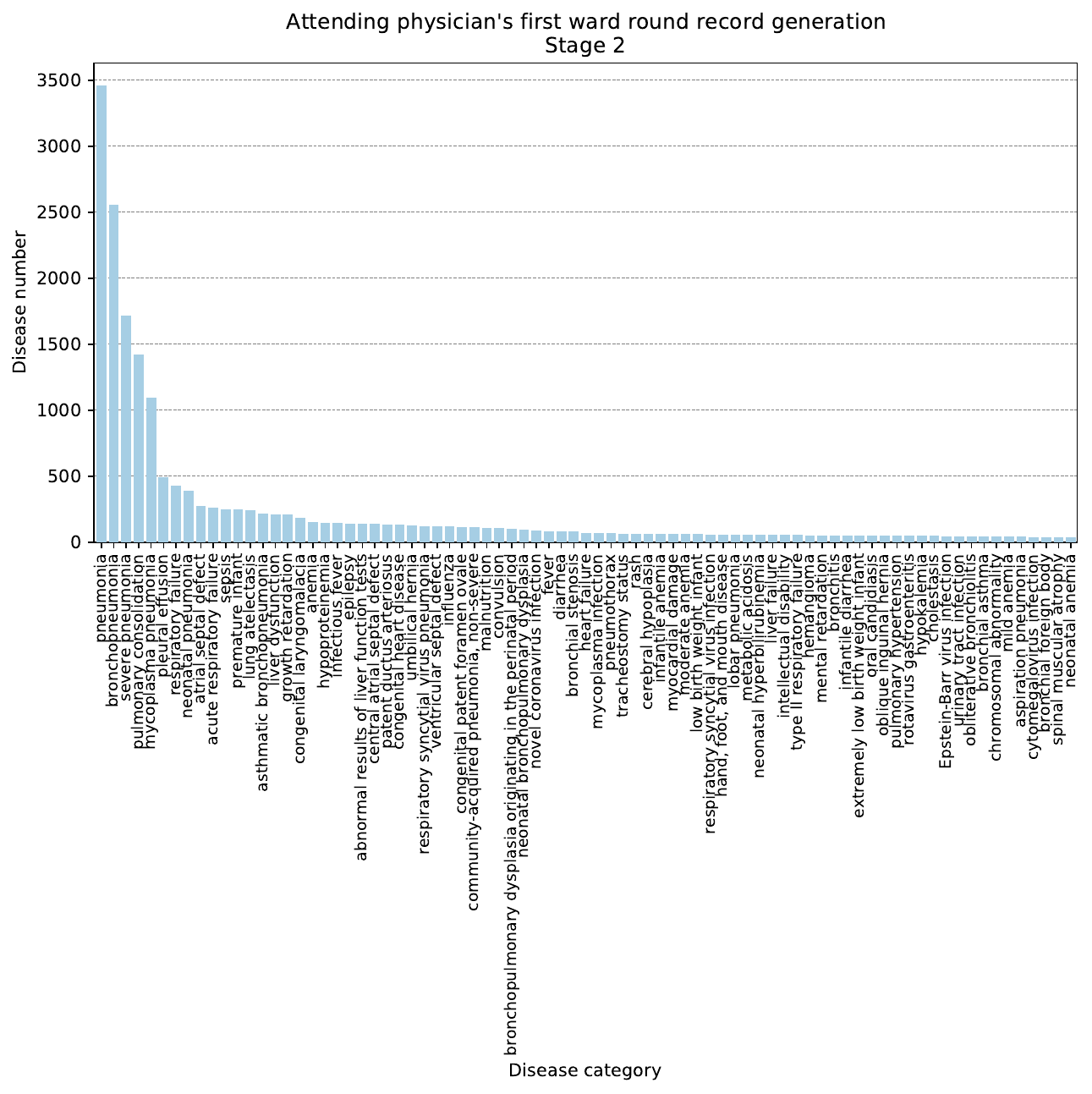}};
      \node[anchor=north west, yshift=4pt] at (image.north west) {\textbf{i}}; 
    \end{tikzpicture}
    \caption{}
    \label{disease_distribution_i}
  \end{subfigure}%
  \hfill
  \begin{subfigure}[b]{0.32\textwidth}
    \begin{tikzpicture}
      \node[anchor=south west,inner sep=0] (image) at (0,0) {\includegraphics[height=4.5cm, width=\linewidth]{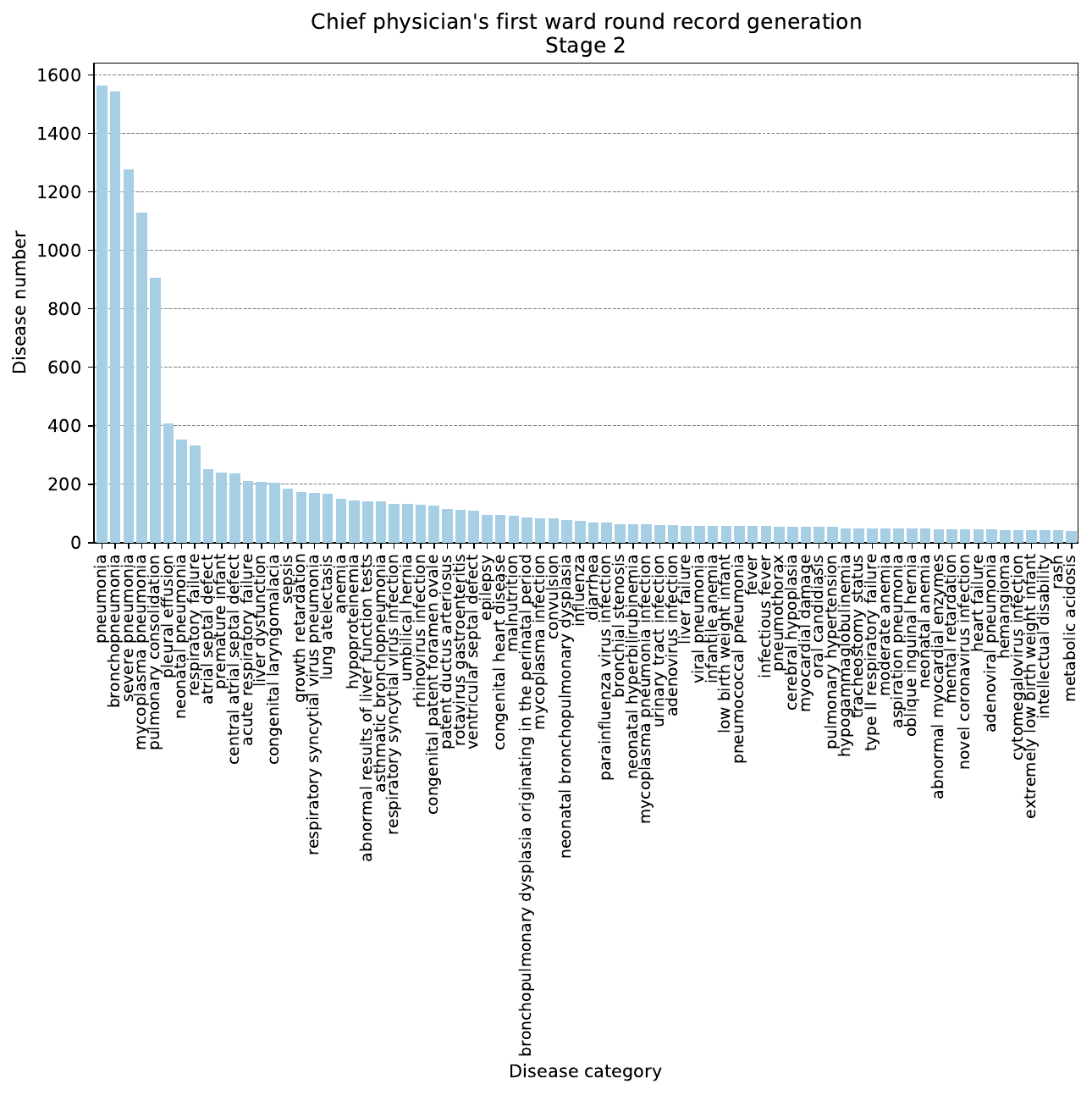}};
      \node[anchor=north west, yshift=4pt] at (image.north west) {\textbf{j}}; 
    \end{tikzpicture}
    \caption{}
    \label{disease_distribution_j}
  \end{subfigure}
  \vspace{-0.8em}
  
  \begin{subfigure}[b]{0.32\textwidth}
    \begin{tikzpicture}
      \node[anchor=south west,inner sep=0] (image) at (0,0) {\includegraphics[height=4.5cm, width=\linewidth]{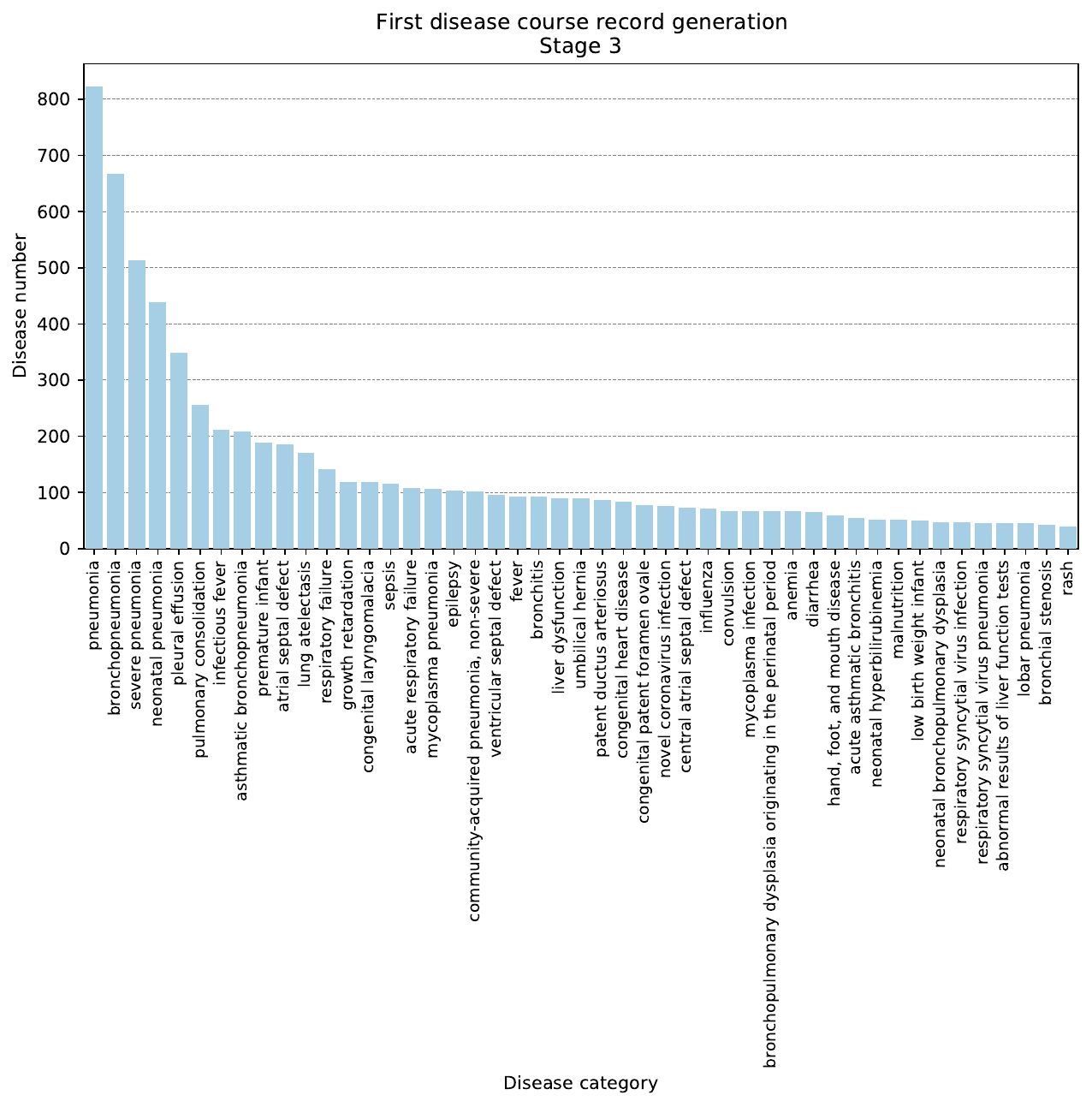}};
      \node[anchor=north west, yshift=4pt] at (image.north west) {\textbf{k}}; 
    \end{tikzpicture}
    \caption{}
    \label{disease_distribution_k}
  \end{subfigure}
  \hfill
  \begin{subfigure}[b]{0.32\textwidth}
    \begin{tikzpicture}
      \node[anchor=south west,inner sep=0] (image) at (0,0) {\includegraphics[height=4.5cm, width=\linewidth]{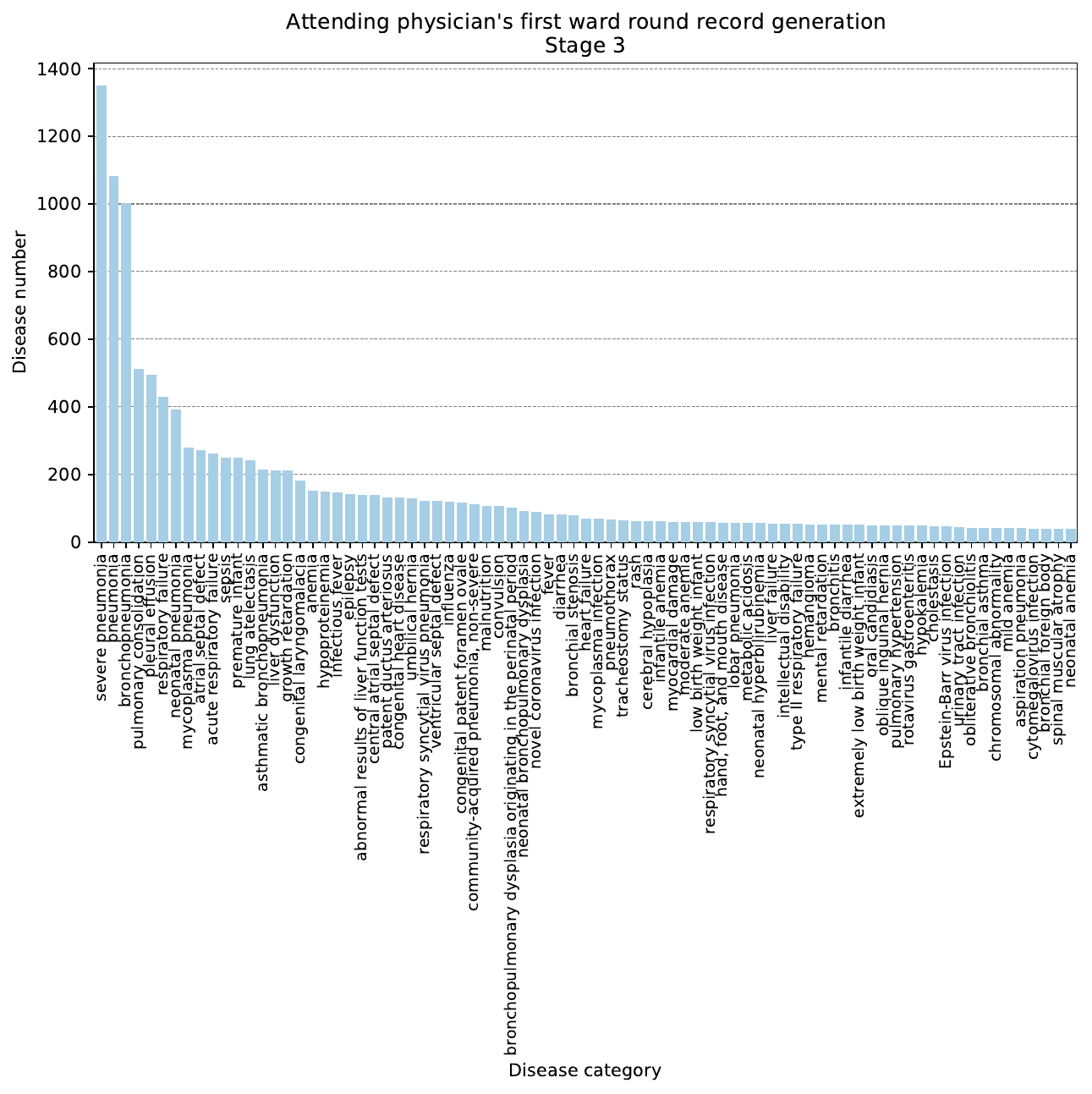}};
      \node[anchor=north west, yshift=4pt] at (image.north west) {\textbf{l}}; 
    \end{tikzpicture}
    \caption{}
    \label{disease_distribution_l}
  \end{subfigure}%
  \hfill
  \begin{subfigure}[b]{0.32\textwidth}
    \begin{tikzpicture}
      \node[anchor=south west,inner sep=0] (image) at (0,0) {\includegraphics[height=4.5cm, width=\linewidth]{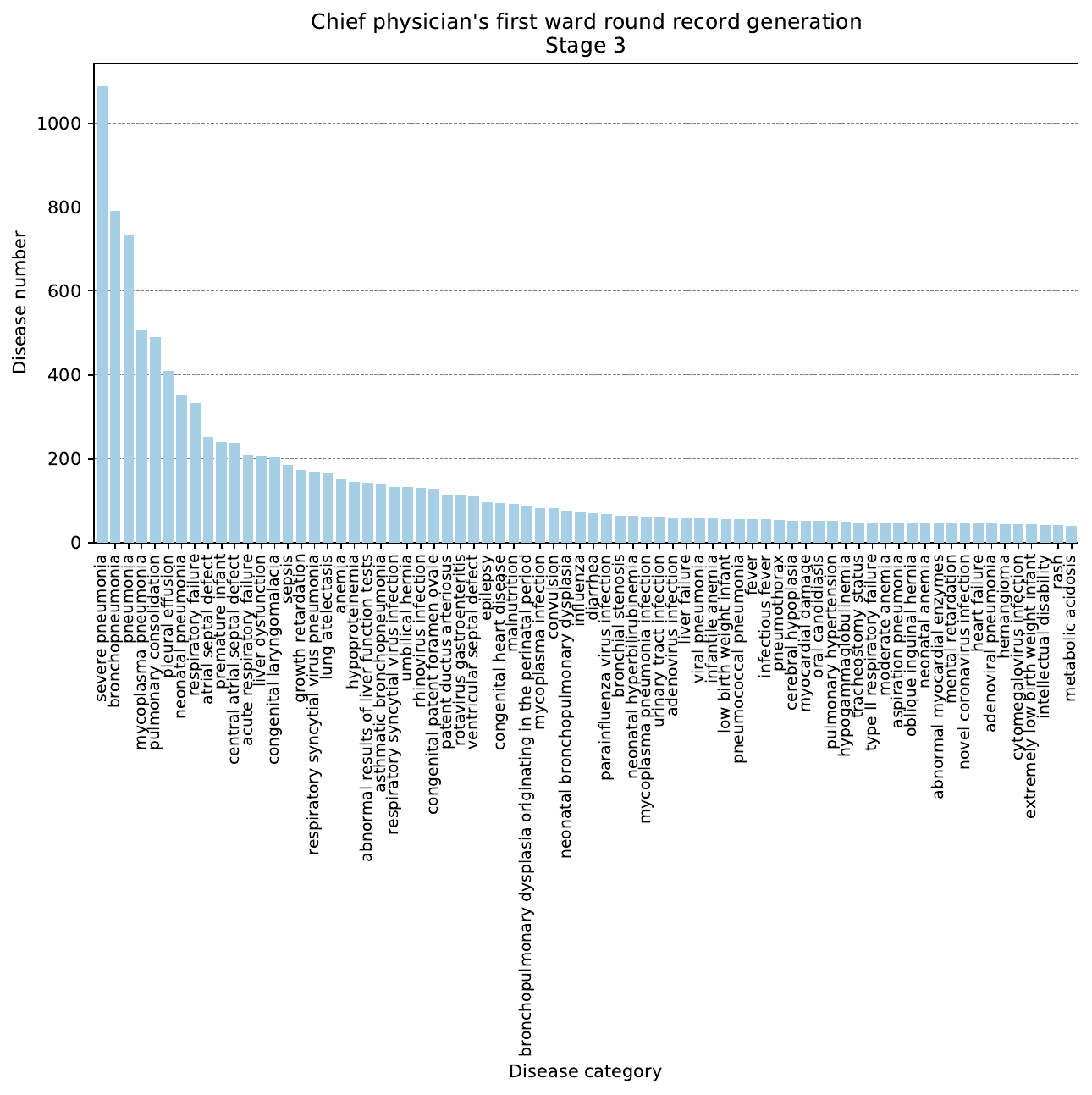}};
      \node[anchor=north west, yshift=4pt] at (image.north west) {\textbf{m}}; 
    \end{tikzpicture}
    \caption{}
    \label{disease_distribution_m}
  \end{subfigure}

  \caption{\textbf{The disease distribution of P2Med-MD across different stages and tasks.} Stage 1 to stage 3 represented medical knowledge infusion pre-training, task type-based balanced instruction-tuning, and disease category-based balanced instruction-tuning, respectively. Note: P2Med-MD: Medical Multimodal Dataset for Pediatric Pneumonia. CT: Computed Tomography.}
  \label{disease_distribution}
\end{figure*}

\subsubsection{Stage 1: Medical Knowledge Infusion Data\label{stage1_data}}
Considering the complexity of inpatient records, stage 1 injected medical knowledge into the general model by learning from all radiology image-report pairs~(including X-ray and CT) and simple outpatient medical records. Fig.~\ref{disease_distribution_a} depicted the disease categories derived from the impression in X-ray radiology reports, with each category comprising over 100 samples during stage 1. Fig.~\ref{disease_distribution_c} displayed the disease categories extracted from the impression in CT radiology reports, each with more than 10 samples. And Fig.~\ref{disease_distribution_d} outlined the disease categories identified from the preliminary diagnosis in outpatient medical records, each encompassing over 400 samples.

However, we observed a significant amount of repetitive descriptions within the outpatient records. Prior researches~\cite{hernandez2022scaling,allal2023santacoder,lee2021deduplicating,penedo2023refinedweb} have demonstrated that repetition in training data can degrade model performance. Thus, it was crucial to perform deduplication to ensure the quality of outpatient records. The data deduplication scheme employed in previous studies~\cite{li2022competition,nijkamp2022codegen,fried2022incoder} typically relied on non-whitespace exact text matching, which was suboptimal due to the diverse writing styles of different doctors. \cite{kocetkov2022stack} indicated that near-deduplication could enhance performance. We followed this pipeline that largely inherited the settings from CodeParrot~\cite{tunstall2022natural}. It involved calculating MinHashes~\cite{broder2000identifying} of all outpatient records and applying Locally Sensitive Hashing~(LSH) to cluster records based on their MinHash fingerprints. During the LSH phase, similar outpatient records were grouped into the same buckets, thereby identifying them as duplicates. From each group of duplicates, only one record was retained. Fig.~\ref{disease_distribution_e} illustrated the disease distribution post-deduplication of these outpatient medical records. 

\subsubsection{Stage 2: Task Type-Based Balanced Sampling Data\label{stage2_data}}
To ensure the model effectively followed diverse task instructions, we curated a variety of instruction-following data covering six distinct tasks in stage 2. Given the two orders of magnitude difference in sample volumes between outpatient and inpatient medical records, it was essential to balance the number of outpatient and inpatient records, as the generative model was sensitive to data imbalances. By looping through disease categories with sample sizes between 325 and 5,000 in outpatient records of stage 1, a maximum of 500 samples per category were sampled without repetition until the total sample size was balanced with that of inpatient records. Due to the multi-label nature of the preliminary diagnosis in outpatient records, the sampled data included a broader range of disease categories. Fig.~\ref{disease_distribution_a},~\ref{disease_distribution_c},~\ref{disease_distribution_f},~\ref{disease_distribution_h},~\ref{disease_distribution_i}, and~\ref{disease_distribution_j} depicted the distribution of disease categories for six tasks during stage 2. Specifically, Fig.~\ref{disease_distribution_h},~\ref{disease_distribution_i}, and~\ref{disease_distribution_j} illustrated the disease categories extracted from the admission diagnosis or current diagnosis in three-level inpatient medical records, each category featuring over 40 samples.

\subsubsection{Stage 3: Disease Category-Based Balanced Sampling Data\label{stage3_data}}
We observed considerable differences in the distribution of disease categories per task in stage 2, potentially impairing the performance of the generative model. To mitigate the long-tail problem, it was essential to perform balanced sampling of disease categories in stage 3. For X-ray image-report pairs, we sampled up to 500 samples per category from those with sample sizes ranging from 100 to 2,000. All CT image-report pairs were included due to the relatively smaller sample size. For outpatient medical records, we sampled up to 200 samples per category from those identified in stage 1 with sample sizes between 325 and 5,000. For three-level inpatient medical records, we included all samples from disease categories containing 40 to 500 samples. Fig.~\ref{disease_distribution_b},~\ref{disease_distribution_c},~\ref{disease_distribution_g},~\ref{disease_distribution_k},~\ref{disease_distribution_l}, and~\ref{disease_distribution_m} illustrated the distribution of disease categories for six tasks in stage 3, which were more balanced compared to stage 2.

\subsubsection{Medical Multimodal Benchmark for Pediatric Pneumonia~(P2Med-MBench)\label{benchmark}}

Building upon P2Med-MD, we presented P2Med-MBench, a comprehensive evaluation benchmark for pediatric pneumonia. P2Med-MBench contained six distinct tasks, including radiology report generation~(X-ray), radiology report generation~(CT), outpatient medical record generation, first disease course record generation, attending physician's first ward round record generation, and chief physician's first ward round record generation. A detailed breakdown of each task, including task description, clinical scenario, modality, image dimension, model input, and model output~\cite{wang2019pediatrics}, was shown in Table \ref{P2Med-MBench}.

\begin{table*}[h]
\caption{\textbf{Description of the P2Med-MBench.}}
\label{P2Med-MBench}
\renewcommand\arraystretch{1.3}
\setlength{\tabcolsep}{4.7pt}
\begin{tabular}{llllllll}
\hline
\begin{tabular}[c]{@{}l@{}}Task \\ number\end{tabular} & \begin{tabular}[c]{@{}l@{}}Task \\ description\end{tabular}                                   & \begin{tabular}[c]{@{}l@{}}Clinical \\ scenario\end{tabular} & Modality                                       & \begin{tabular}[c]{@{}l@{}}Image \\ modality\end{tabular} & \begin{tabular}[c]{@{}l@{}}Image \\ dimension\end{tabular} & \begin{tabular}[c]{@{}l@{}}Model \\ input\end{tabular}                                                                                        & \begin{tabular}[c]{@{}l@{}}Model \\ output\end{tabular}                                                                             \\ \hline
1                                                               & \begin{tabular}[c]{@{}l@{}}Radiology report \\ generation~(X-ray)\end{tabular}                                 & \begin{tabular}[c]{@{}l@{}}Outpatient / \\ Inpatient\end{tabular}     & Image-text & X-ray                                                              & 2D                                                                  & \begin{tabular}[c]{@{}l@{}}Examination time + \\ Examination modality + \\ Image\end{tabular}                                                          & Findings + Impression                                                                                                                         \\ \hline
2                                                               & \begin{tabular}[c]{@{}l@{}}Radiology report \\ generation~(CT)\end{tabular}                                 & \begin{tabular}[c]{@{}l@{}}Outpatient / \\ Inpatient\end{tabular}     & Image-text & CT                                                                 & 3D                                                                  & \begin{tabular}[c]{@{}l@{}}Examination time + \\ Examination modality + \\ Image\end{tabular}                                                          & Findings + Impression                                                                                                                         \\ \hline
3                                                               & \begin{tabular}[c]{@{}l@{}}Outpatient medical \\ record generation\end{tabular}                        & Outpatient                                                            & Plain text                                                    & -                                                                  & -                                                                   & \begin{tabular}[c]{@{}l@{}}Chief complaint + \\ History of present illness + \\ Physical examination\end{tabular}                                      & \begin{tabular}[c]{@{}l@{}}Preliminary diagnosis + \\ Treatment recommendation + \\ Treatment plan\end{tabular}                             \\ \hline
4                                                               & \begin{tabular}[c]{@{}l@{}}First disease course \\ record generation\end{tabular}                      & Inpatient                                                             & Plain text                                                    & -                                                                  & -                                                                   & \begin{tabular}[c]{@{}l@{}}History of present illness + \\ Physical examination + \\ Auxiliary examination + \\ Clinical history features\end{tabular} & \begin{tabular}[c]{@{}l@{}}Diagnostic basis + \\ Admission diagnosis + \\ Diagnostic and treatment plan\end{tabular}                         \\ \hline
5                                                               & \begin{tabular}[c]{@{}l@{}}Attending physician's \\ first ward round \\ record generation\end{tabular} & Inpatient                                                             & Plain text                                                    & -                                                                  & -                                                                   & \begin{tabular}[c]{@{}l@{}}Clinical history features + \\ Additional clinical history and signs\end{tabular}                                           & \begin{tabular}[c]{@{}l@{}}Diagnostic basis + \\ Current diagnosis + \\ Diagnostic and treatment plan\end{tabular} \\ \hline
6                                                               & \begin{tabular}[c]{@{}l@{}}Chief physician's \\ first ward round \\ record generation\end{tabular}     & Inpatient                                                             & Plain text                                                    & -                                                                  & -                                                                   & \begin{tabular}[c]{@{}l@{}}Clinical history features + \\ Additional clinical history and signs\end{tabular}                                           & \begin{tabular}[c]{@{}l@{}}Diagnostic basis + \\ Current diagnosis + \\ Diagnostic and treatment plan\end{tabular} \\ \hline
\end{tabular}
\begin{tablenotes}[para,flushleft]
\item Note: P2Med-MBench: Medical Multimodal Benchmark for Pediatric Pneumonia. CT: Computed Tomography.
\end{tablenotes}
\end{table*}

It was noteworthy that the P2Med-MD was collected over a continuous period. It was diverse and complex, potentially even containing some data with noise. To guarantee the data quality and representativeness for evaluation, two pediatric pulmonology specialists performed meticulous manual verification of the P2Med-MBench samples. Ultimately, we obtained 121 samples for radiology report generation~(X-ray), 121 samples for radiology report generation~(CT), 100 samples for outpatient medical record generation, 100 samples for first disease course record generation, 100 samples for attending physician's first ward round record generation, and 100 samples for chief physician's first ward round record generation. These samples were dismissed in the whole training set. Detailed descriptions of the six evaluation tasks were provided in the following.

\textbf{Radiology report generation~(X-ray).}
This task primarily focused on the automatic generation of radiology reports for X-ray images, encompassing two key sections: findings and impression. The former provided a detailed description of crucial aspects observed in the 2D X-ray images, while the latter summarized the most relevant findings. Given that an outpatient or inpatient might have one or more X-ray images taken from various views and different times, we incorporated time, modality, and corresponding multi-view images in the input to facilitate correlation and comparison with prior radiological data of the same patient, thereby enabling the generation of more objective and comprehensive radiology reports. For a given set of X-ray images, we employed prompt sentences similar to the following as input: ``\textit{Current radiological data is as follows: \textbackslash{}n [Examination time] December 31, 2022 \textbackslash{}n [Examination modality] X-ray \textbackslash{}n [Image] \textless{}image\textgreater{}...\textless{}image\textgreater{} \textbackslash{}n Based on the above information, combined with professional radiological knowledge, generate a report in the format: \textbackslash{}n [Findings] \{Your findings based on the images\} \textbackslash{}n [Impression] \{Your impression based on the images\} \textbackslash{}n}". The number of \textless{}image\textgreater{} tokens corresponded to the number of views, with one for the anteroposterior view and two for the anteroposterior and lateral views. The impression, as the most critical component, was assessed using two metrics: accuracy and comprehensiveness, while the findings were evaluated solely on accuracy. To ensure the reliability of our evaluation, we have selected 100 sets of X-ray image-report pairs at unique time points and 10 sets at multiple times from the same patient, altogether comprising 121 samples and covering more than 47 distinct diseases.

\textbf{Radiology report generation~(CT).}
This task was similar to the radiology report generation~(X-ray) task but was specifically designed for 3D CT images, thereby the examination modality was CT. The number of \textless{}image\textgreater{} tokens denoted the number of series, with one representing a non-contrast series, and two indicating both non-contrast and contrast-enhanced series. The final selection of 121 samples encompassed more than eight types of diseases.

\textbf{Outpatient medical record generation.}
This task imitated the clinical process of a physician's outpatient visit, utilizing textual information such as the chief complaint, history of present illness, and physical examination to formulate a preliminary diagnosis, a treatment recommendation for the patient, and a treatment plan for the doctor. Here, we simulated this task as a prompt-based generative dialogue task. For example, we used the following as input: ``\textit{Current outpatient pediatric information is as follows: \textbackslash{}n [Chief complaint] The pediatric patient presented with cough and fever for 4 days ... \textbackslash{}n [History of present illness] Maximum temperature of 40$^{\circ}$C ... \textbackslash{}n [Physical examination] The pediatric patient is conscious and responsive ... \textbackslash{}n Based on the above information, combined with professional medical knowledge, make a diagnosis in the format: \textbackslash{}n [Preliminary diagnosis] \{Your preliminary diagnosis\} \textbackslash{}n [Treatment recommendation] \{Your treatment recommendation\}\textbackslash{}n [Treatment plan] \{Your treatment plan\} \textbackslash{}n}". The output was then matched with the ground truth. The preliminary diagnosis, the most crucial element, was assessed on both accuracy and comprehensiveness, while the treatment recommendation and treatment plan were evaluated only on accuracy. We have selected 100 samples covering more than 55 disease types for preliminary diagnosis.

\textbf{First disease course record generation.}
This task simulated the process by which a resident physician recorded the first disease course for a patient within 24 hours of hospitalization, synthesizing textual information such as history of present illness, physical examination, auxiliary examination, and clinical history features to predict the diagnostic basis, admission diagnosis, and diagnostic and treatment plan. The diagnostic basis explained the causes related to the admission diagnosis, reflecting the model's capacity for logical reasoning. We employed prompt sentences like ``\textit{Current inpatient pediatric information is as follows: \textbackslash{}n [History of present illness] The pediatric patient had a fever without obvious inducement five days ago~(October 9, 2023) ... \textbackslash{}n [Physical examination] The pediatric patient is conscious and responsive ... \textbackslash{}n [Auxiliary examination] October 9, 2023: outpatient blood test ... \textbackslash{}n [Clinical history features] Male, 13 years old ... \textbackslash{}n Based on the above information, combined with professional medical knowledge, make a diagnosis in the format: \textbackslash{}n [Diagnostic basis] \{Your diagnostic basis\} \textbackslash{}n [Admission diagnosis] \{Your admission diagnosis\} \textbackslash{}n [Diagnostic and treatment plan] \{Your diagnostic and treatment plan\} \textbackslash{}n}" as input. We focused primarily on the admission diagnosis, evaluating it for both accuracy and comprehensiveness, while the diagnostic basis and diagnostic and treatment plan were assessed only for accuracy. Similarly, we have selected 100 samples that include more than 47 types of diseases for admission diagnosis.

\textbf{Attending physician's first ward round record generation.}
This task simulated how an attending physician performed the first ward round within 72 hours of a patient's hospitalization, analyzing textual data such as clinical history features and additional clinical history and signs to predict the diagnostic basis, current diagnosis, and diagnostic and treatment plan. We utilized the following prompt as input: ``\textit{Current inpatient pediatric information is as follows: \textbackslash{}n [Clinical history features] Male, 13 years old ... \textbackslash{}n [Additional clinical history and signs] The pediatric patient continues to experience recurrent fever, peaking at 39.7$^{\circ}$C ... \textbackslash{}n Based on the above information, combined with professional medical knowledge, make a diagnosis in the format: \textbackslash{}n [Diagnostic basis] \{Your diagnostic basis\} \textbackslash{}n [Current diagnosis] \{Your current diagnosis\} \textbackslash{}n [Diagnostic and treatment plan] \{Your diagnostic and treatment plan\} \textbackslash{}n}". Our primary focus was on the current diagnosis, thus we evaluated the prediction using accuracy and comprehensiveness, while the diagnostic basis and the diagnostic and treatment plan were assessed solely on accuracy. Similarly, we have selected 100 samples, covering over 79 types of diseases for current diagnosis.

\textbf{Chief physician's first ward round record generation.}
This task was similar to the attending physician's first ward round record generation task; however, it specifically simulated the chief physician's first ward round record within one week of a patient's hospitalization. The selected 100 samples encompassed more than 74 types of diseases for current diagnosis.

\begin{figure*}[htbp]
\centerline{\includegraphics[width=\textwidth]{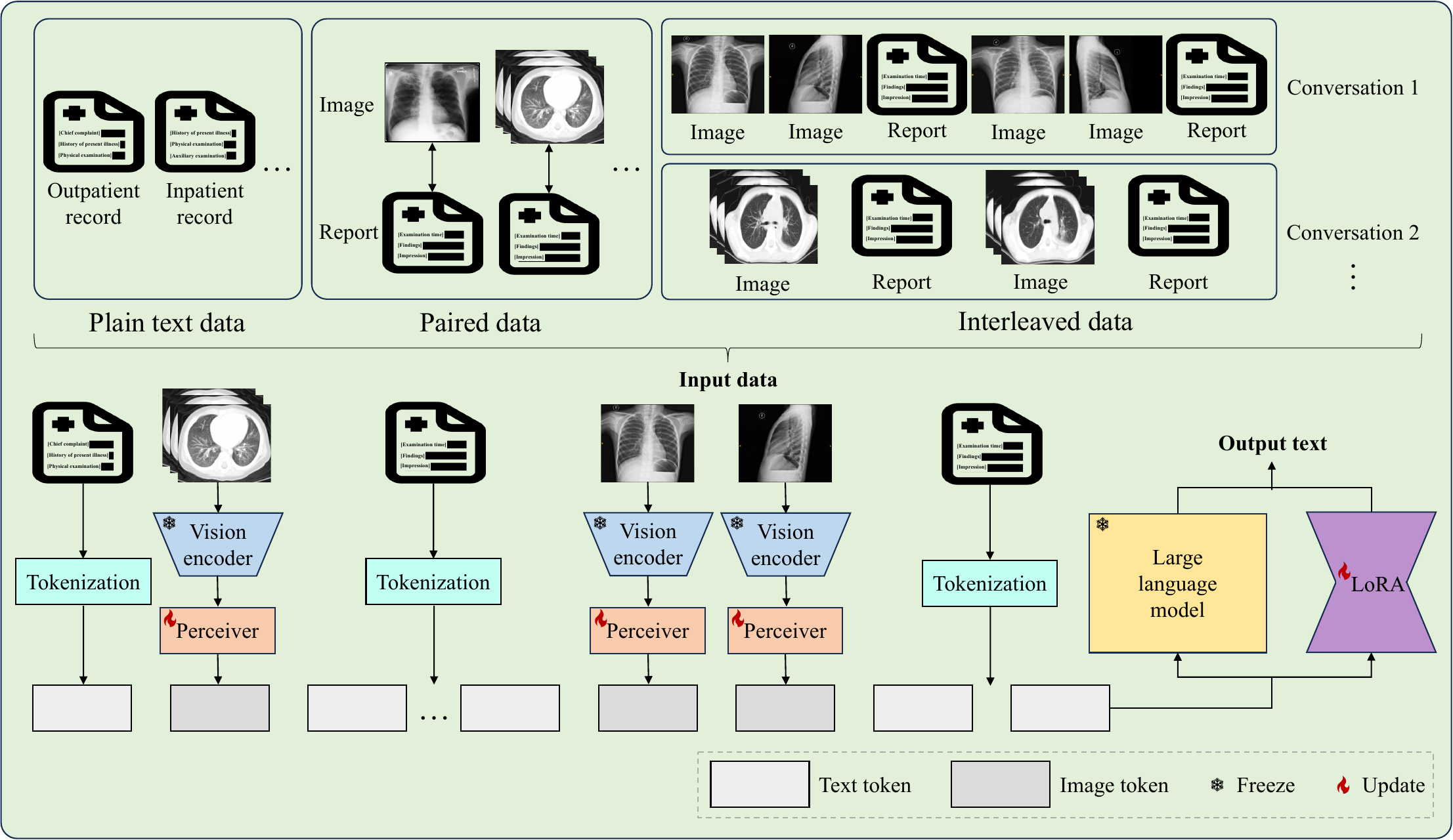}}
\vspace{-0.5em}
\caption{\textbf{The framework of our P2Med-MLLM.} It can process plain text, medical image-report pairs, and multiple 2D or 3D images interleaved with medical reports. Note: P2Med-MLLM: Medical Multimodal Large Language Model for Pediatric Pneumonia.}
\label{framework}
\vspace{-1.5em}
\end{figure*}

\subsection{Medical Multimodal Large Language Model for Pediatric Pneumonia~(P2Med-MLLM)\label{model}}
As illustrated in Fig.~\ref{framework}, the architecture of the \textbf{Med}ical \textbf{M}ultimodal \textbf{L}arge \textbf{L}anguage \textbf{M}odel for \textbf{P}ediatric \textbf{P}neumonia~(P2Med-MLLM) primarily consisted of three modules: a pre-trained LLM~(\textit{e.g.}, Chinese-LLaMA-2~\cite{cui2023efficient,touvron2023llama}) serving as the foundational model, a pre-trained vision encoder~(\textit{e.g.}, CLIP~\cite{radford2021learning}) responsible for encoding medical images into image embeddings, and an attention-based perceiver~\cite{alayrac2022flamingo} that transformed these image embeddings into image tokens compatible with the LLM.

For P2Med-MLLM training, we considered a three-stage procedure, shown in Fig.~\ref{overview_flowchart}. During the first stage of medical knowledge fusion pre-training, only the perceiver module was trainable, facilitating the alignment of multimodal features. Subsequently, during the second and third stages of instruction-tuning, the LLM employed Low-Rank Adaptation~(LoRA)~\cite{hu2021lora} for efficient parameter tuning. All medical data were formatted as either a single-round conversation~(plain text and image-text paired data) or multi-round conversations~(interleaved image-text data) for model training. Next, we would provide a detailed introduction to the P2Med-MLLM.

\subsubsection{Efficient Large Language Model Finetuning\label{stage1_model}} 
The LLM pre-trained on web datasets lacked the vertical domain knowledge required for pediatric pneumonia, leading to suboptimal performance for corresponding medical tasks. It was essential to update the LLM parameters using medical data. Due to constraints in computational resources, finetuning the full parameters of the LLM, which consisted of 7B parameters, was unfeasible. To address these challenges, we adopted LoRA for efficient parameter tuning.

LoRA introduced low-rank matrices, denoted as \(A\) and \(B\), which had a significantly smaller number of parameters than the original model weights. The adaptation was formulated as:
 
\begin{equation}
W' = W + \Delta W
\end{equation}
where \( W \) was the original weight matrix, and \( \Delta W \) was the low-rank update defined as:
\begin{equation}
\Delta W = A B^T
\end{equation}
where \( A \in \mathbb{R}^{d \times r} \) and \( B \in \mathbb{R}^{d \times r} \), with \( r \) being the rank which was much smaller than \( d \), the dimension of \( W \). \( T \) represented the transpose operation.

By training only the low-rank matrices \( A \) and \( B \) while keeping the original LLM parameters frozen, we achieved efficient optimization with significantly reduced computational overhead. The lightweight nature of these low-rank matrices ensured that there was almost no additional inference latency introduced during the inference stage. Ultimately, we efficiently incorporated critical medical knowledge into the LLM, enhancing its performance in this specialized field without the need for extensive computational resources.

\subsubsection{2D/3D Medical Image Perception\label{stage2_model}}
Traditional approaches typically employed a linear projection~\cite{liu2024visual,li2024llava} or a Multi-Layer Perceptron~(MLP)~\cite{liu2024improved} as the cross-modal connector to convert medical image embeddings into visual tokens for being integrated into LLM. However, these conventional methods encountered significant challenges when processing 3D CT images, which typically consisted of more than 30 slices. The conversion of these images resulted in an excessively large number of visual tokens, far exceeding the LLM's maximum token limit. For instance, a 2D image was encoded as 576 visual tokens, whereas a 3D image with 30 slices was encoded as \(30 \times 576 = 17,280\) visual tokens, which far exceeded the typical LLM maximum length of 4,096 tokens.

To address this challenge, we utilized a lightweight Transformer decoder-only~\cite{vaswani2017attention} structure based on the attention mechanism, named as the perceiver module~\cite{alayrac2022flamingo,li2023blip}, to simultaneously process 2D/3D medical images embeddings into a fixed number of visual tokens. Specifically, the perceiver first incorporated learnable temporal and positional embeddings into the image embeddings, which were then flattened for injecting into the attention layer. The attention layer operates were as follows:

\begin{equation}
Q = W^Q h,  K = W^K x, V = W^V x
\end{equation}
where \( h \) represented the learnable latent array, while \( x \) corresponded to the flattened visual features. \( Q \), \( K \), and \( V \) denoted the query, key, and value vectors used in cross-attention interactions. \( W^Q \), \( W^K \), and \( W^V \) were learned weight matrices.

The attention mechanism then computed:
\begin{equation}
\text{Attention}(Q, K, V) = \text{softmax}\left(\frac{QK^T}{\sqrt{d_k}}\right)V
\end{equation}
where \( d_k \) was the dimension of the key vector.

Ultimately, through this unified architecture, the perceiver module efficiently processed the perception of both 2D and 3D medical images, ensuring that the resulting visual tokens were compatible in length with the LLM. This approach optimized the integration of complex medical imaging data with the LLM, making it suitable for advanced medical tasks.

\subsubsection{Multimodal Medical Data Formats\label{stage3_model}}
The medical data collected for training were categorized into plain text data and multimodal data. Each data instance input \(X_i\) and output \(X_o\) were reformatted into an instruction-following structure: 
 \[ \texttt{Human:} \; \mathit{X}_p \; \mathit{X}_i^{1} \; \textcolor{lightgreen}{\langle \texttt{STOP} \rangle} \; \texttt{Assistant:} \; \textcolor{lightgreen}{\mathit{X}_o^{1} \; \langle \texttt{STOP} \rangle} \]
where \(X_p\) referred predefined instructional prompts for different tasks. Please see Sec~\ref{benchmark} for the illustrations of different prompts. The model was designed to predict the assistant's responses and where to stop. Therefore, only \textcolor{lightgreen}{green tokens} were used to calculate the training loss.

For the plain text data, which primarily concerned record generation, \(X_i\) referred to patient-specific information, while \(X_o\) consisted of the resultant medical records. In the context of multimodal medical image-report data, the input \(X_i\) referred to visual data such as X-ray or CT images, and \(X_o\) consisted of the findings and impressions, interpreting and summarizing the visual observations. Qualitative examples illustrating our data formats were shown in Fig.~\ref{qualitative_results_english}. For the original Chinese version, please refer to Fig.~\ref{qualitative_results_chinese}.

The analysis of the collected data revealed that a patient sometimes underwent multiple radiological examinations, resulting in correlated medical image-report pairs. For instance, sequential reports may contain comparisons such as ``\textit{Compared to the scan from October 29, 2022, both lungs show ...}". Treating each medical image-report pair as an independent instruction instance can hinder the model's ability to recognize relationships across a patient's sequential image-report data. To mitigate this limitation, we converted multiple related image-report pairs of a patient into an interleaved data format:
\[
\begin{aligned}
&\texttt{Human:} \; \mathit{X}_p \; \mathit{X}_i^{1} \; \textcolor{lightgreen}{\langle \texttt{STOP} \rangle} \; \texttt{Assistant:} \; \textcolor{lightgreen}{\mathit{X}_o^{1} \; \langle \texttt{STOP} \rangle}, \\
&\texttt{Human:} \; \mathit{X}_p \; \mathit{X}_i^{2} \; \textcolor{lightgreen}{\langle \texttt{STOP} \rangle} \; \texttt{Assistant:} \; \textcolor{lightgreen}{\mathit{X}_o^{2} \; \langle \texttt{STOP} \rangle} \ldots
\end{aligned}
\]

This approach enabled our model to consider all previously associated image-report data when generating new reports.

\subsection{Training Details}

\subsubsection{Data Preprocessing}
In the analysis of medical imaging and textual data for pediatric pneumonia, we initially de-identified all patient-related information. For the preprocessing of 2D chest X-ray images, each chest X-ray examination retained either an anteroposterior view or both anteroposterior and lateral views, and the x-axis and y-axis were resized to 336 pixels. For the preprocessing of 3D chest CT images, we selected lung reconstruction series with a slice thickness of 5.0 mm and normalized them based on a window level of -500 HU and a window width of 1,200 HU. Each chest CT examination retained either a non-contrast series or both non-contrast and contrast-enhanced series, and the x-axis and y-axis were resized to 336 pixels. When the z-axis dimensions of non-contrast and contrast-enhanced series differed, the shorter one was padded with zeros to match. For the preprocessing of textual data including medical reports, outpatient, and inpatient records, we removed records with any missing element in the model's ground truth output. Additionally, we excluded records containing over 4,000 tokens, as their excessive lengths hindered the effective learning of the LLM. Specifically, for deduplication of outpatient records, we utilized 5-grams with a Jaccard similarity threshold of 0.85, 16 rows, and 256 bands. Meanwhile, we filtered out outpatient records with n-grams less than 5.

\subsubsection{Implementation}
We utilized a 24-layer, 2D ViT-L/14 with 1,024 embedding dimensions as the vision encoder, initialized with CLIP weights. The perceiver was a 6-layer transformer decoder with a learnable latent array of 32 $\times$ 4,096 dimensions. For the LLM, we employed the 32-layer, 7B Chinese-LLaMA-2. Our final model comprised 8B parameters. During the three training stages, we froze the vision encoder and LLM, updating only the perceiver and LoRA parameters. All models were implemented in PyTorch and trained on 8 NVIDIA A6000 GPUs with 48 GB memory each. To prevent gradient errors during backpropagation, each batch during training samples was either image-text pairs or plain text data. For optimization, we used the Adam optimizer with a cosine decay scheduler and a warmup ratio of 0.03. Detailed hyperparameters were provided in Table \ref{hyperparameters}.

\begin{table}[h]
\begin{threeparttable}
\caption{\textbf{P2Med-MLLM hyperparameters.} Stage 1 to stage 3 represented medical knowledge infusion pre-training, task type-based balanced instruction-tuning, and disease category-based balanced instruction-tuning, respectively.}
\label{hyperparameters}
\renewcommand\arraystretch{1.5}
\setlength{\tabcolsep}{4.0pt}
\begin{tabular}{lllllll}
\hline
Stage & Epoch & \begin{tabular}[c]{@{}l@{}}Training \\ time\end{tabular} & \begin{tabular}[c]{@{}l@{}}Max \\ tokens\end{tabular} & \begin{tabular}[c]{@{}l@{}}Learning \\ rate\end{tabular} & \begin{tabular}[c]{@{}l@{}}Batch size \\ per device\end{tabular} & \begin{tabular}[c]{@{}l@{}}Gradient \\ accumulation\end{tabular} \\ \hline
1     & 1     & 15 hours                                                 & 2,048                                                  & $2 \times 10^{-6}$                                                & 2                                                                & 1                                                                \\ \hline
2     & 4     & 19 hours                                                 & 4,096                                                  & $2 \times 10^{-6}$                                                 & 1                                                                & 16                                                               \\ \hline
3     & 4     & 8 hours                                                  & 4,096                                                  & $2 \times 10^{-7}$                                                 & 1                                                                & 16                                                               \\ \hline
\end{tabular}
\begin{tablenotes}[para,flushleft]
\item Note: P2Med-MLLM: Medical Multimodal Large Language Model for Pediatric Pneumonia.
\end{tablenotes}
\end{threeparttable}
\end{table}

\bibliographystyle{IEEEtran}
\bibliography{tmi}

\clearpage
\onecolumn
\renewcommand{\thefigure}{S\arabic{figure}}
\setcounter{figure}{0}

\section*{Supplementary Materials}

\begin{figure*}[h]
\centerline{\includegraphics[width=\textwidth]{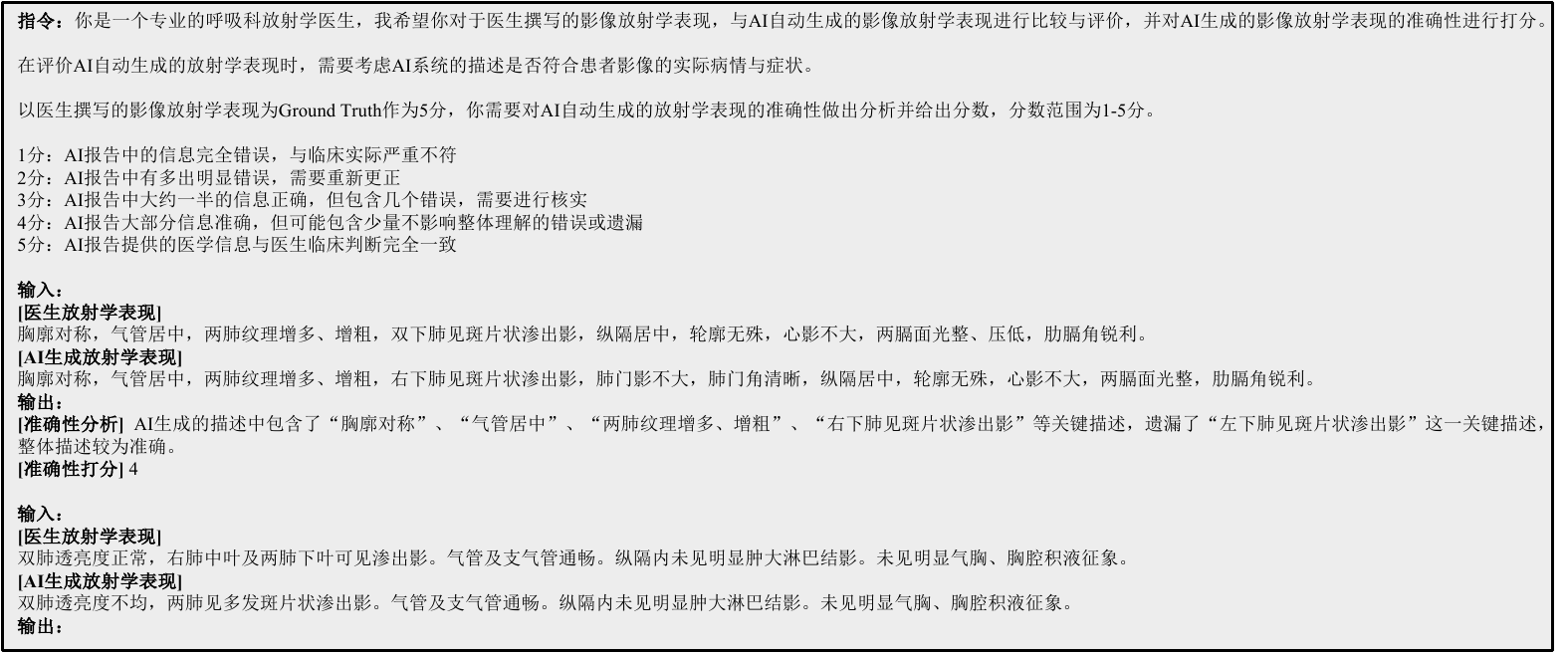}}
\caption{\textbf{Illustration of the evaluation process~(Chinese version).} We evaluated model-generated answers using 13B Chinese-LLaMA-2.}
\label{automatic_rating_chinese}
\end{figure*}

\begin{figure*}[h]
\centerline{\includegraphics[width=\textwidth]{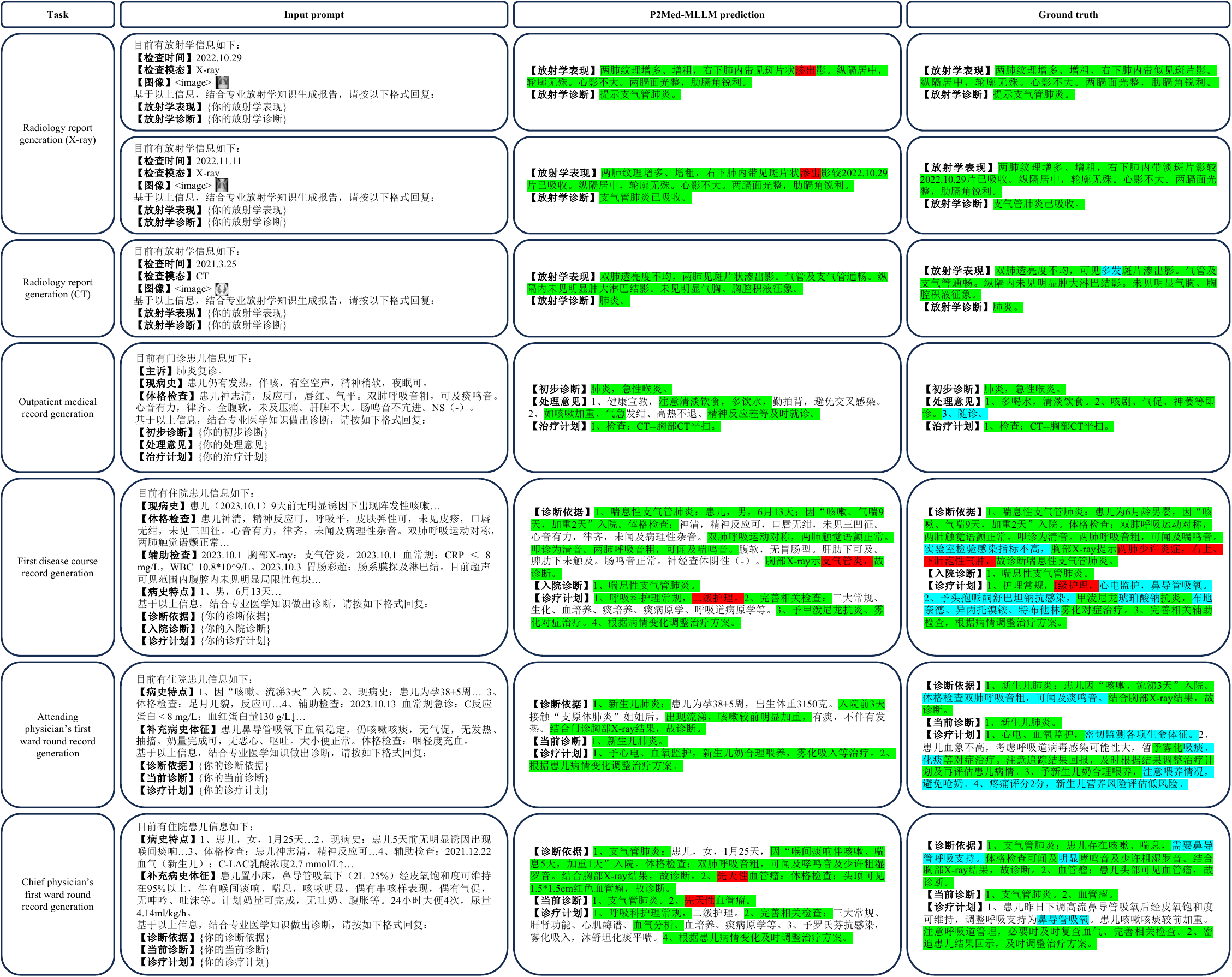}}
\caption{\textbf{Qualitative examples of six different evaluation tasks~(Chinese version).} We presented input prompts along with answers generated by P2Med-MLLM and the target ground truth. The green color in the figure highlighted correct predictions, the red color indicated errors, and the blue color denoted neglected parts. Note: P2Med-MLLM: Medical Multimodal Large Language Model for Pediatric Pneumonia. CT: Computed Tomography.}
\label{qualitative_results_chinese}
\end{figure*}

\end{document}